%% file: main.tex
\documentclass[sigconf, nonacm]{acmart}
\usepackage{amsmath}
\usepackage{amsthm}
\usepackage{subfig}
\usepackage{framed}
\usepackage{graphicx}
\usepackage{color}
\usepackage{multirow}
\usepackage[noend]{algorithmic}
\usepackage{algorithm}

\newtheorem{definition}{Definition}

\DeclareMathOperator*{\argmin}{arg\,min}

\AtBeginDocument{%
  \providecommand\BibTeX{{%
    \normalfont B\kern-0.5em{\scshape i\kern-0.25em b}\kern-0.8em\TeX}}}

\usepackage{todonotes}

\begin{document}
\title{Spatial Knowledge-Infused Hierarchical Learning: An Application in Flood Mapping on Earth Imagery}

\author{Zelin Xu}
\email{zelin.xu@ufl.edu}
\affiliation{\institution{Department of CISE\\University of Florida}
\city{Gainesville}
\state{Florida}
\country{USA}
}

\author{Tingsong Xiao}
\email{xiaotingsong@ufl.edu}
\affiliation{\institution{Department of CISE\\University of Florida}
\city{Gainesville}
\state{Florida}
\country{USA}
}

\author{Wenchong He}
\email{whe2@ufl.edu}
\affiliation{\institution{Department of CISE\\University of Florida}
\city{Gainesville}
\state{Florida}
\country{USA}
}

\author{Yu Wang}
\email{yuwang1@ufl.edu}
\affiliation{\institution{Department of MAE\\University of Florida}
\city{Gainesville}
\state{Florida}
\country{USA}
}

\author{Zhe Jiang}
\authornote{Corresponding author.}
\email{zhe.jiang@ufl.edu}
\affiliation{\institution{Department of CISE\\University of Florida}
\city{Gainesville}
\state{Florida}
\country{USA}
}

\renewcommand{\shortauthors}{Zelin Xu et al.}


\begin{abstract}
Deep learning for Earth imagery plays an increasingly important role in geoscience applications such as agriculture, ecology, and natural disaster management. Still, progress is often hindered by the limited training labels. Given Earth imagery with limited training labels, a base deep neural network model, and a spatial knowledge base with label constraints, our problem is to infer the full labels while training the neural network.
The problem is challenging due to the sparse and noisy input labels, spatial uncertainty within the label inference process, and high computational costs associated with a large number of sample locations. Existing works on neuro-symbolic models focus on integrating symbolic logic into neural networks (e.g., loss function, model architecture, and training label augmentation), but these methods do not fully address the challenges of spatial data (e.g., spatial uncertainty, the trade-off between spatial granularity and computational costs).  
To bridge this gap, we propose a novel Spatial Knowledge-Infused Hierarchical Learning (SKI-HL) framework that iteratively infers sample labels within a multi-resolution hierarchy. Our framework consists of a module to selectively infer labels in different resolutions based on spatial uncertainty and a module to train neural network parameters with uncertainty-aware multi-instance learning. 
Extensive experiments on real-world flood mapping datasets show that the proposed model outperforms several baseline methods.
The code is available at \url{https://github.com/ZelinXu2000/SKI-HL}.
\end{abstract}




\maketitle

\input{intro}

\input{problem}

\input{approach}

\input{evaluation}

\section{Conclusion and Future Works}
In this paper, we proposed a novel Spatial Knowledge-Infused Hierarchical Learning (SKI-HL) framework that successfully addresses the limitations of existing neuro-symbolic models, particularly with regard to spatial data, through a system of iteratively inferring labels within a multi-resolution hierarchy. Our model outperformed several baseline methods on real-world flood mapping datasets. 

However, there is still a vast scope for further exploration and improvement in various aspects of the work. First, the model can incorporate temporal dynamics features and capture changes in earth imagery over time, which is critical for many applications such as deforestation tracking. Second, we can expand to other geospatial applications to further validate the generalizability and adaptability of the proposed SKI-HL framework.

\section*{Acknowledgement}
This material is based upon work supported by the National Science Foundation (NSF) under Grant No. IIS-2147908, IIS-2207072, CNS-1951974, OAC-2152085, and the National Oceanic and Atmospheric Administration grant NA19NES4320002 (CISESS) at the University of Maryland.

\bibliographystyle{ACM-Reference-Format}
{\bibliography{refs}}
\end{document}

%% file: intro.tex
\section{Introduction}
Deep learning for spatial data (e.g., Earth imagery) plays an important role in many applications such as transportation \cite{gao2019incorporating, fabrizio2006knowledge}, agriculture \cite{russwurm2017temporal, qiao2023kstage}, ecology \cite{harmon2022injecting}, urban planning \cite{jean2019tile2vec}, and natural hazards management \cite{sainju2020hidden, he2022earth}. Unfortunately, one major bottleneck is that deep learning models require a large number of training labels (e.g., ImageNet), which is often unavailable in geoscience domains \cite{karpatne2022knowledge}. 
This paper studies the integration of spatial domain knowledge and deep learning to overcome limited training labels. 

\begin{figure}[h]
    \centering
    \subfloat[Map and labeled location]{
    \includegraphics[width=0.125\textwidth]{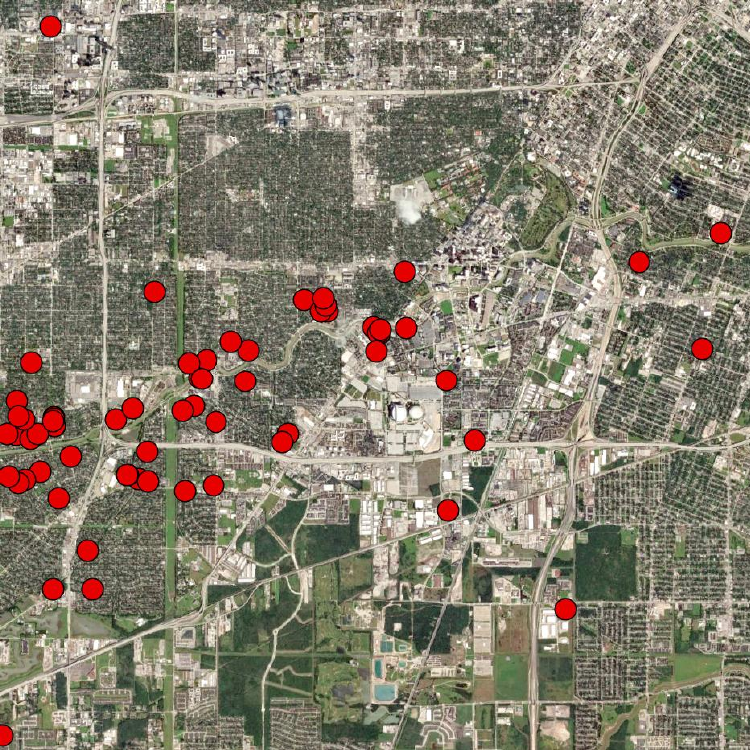}
    \label{fig:problem_map}
    }   
  \subfloat[Logic inference]{
    \includegraphics[width=0.3\textwidth]{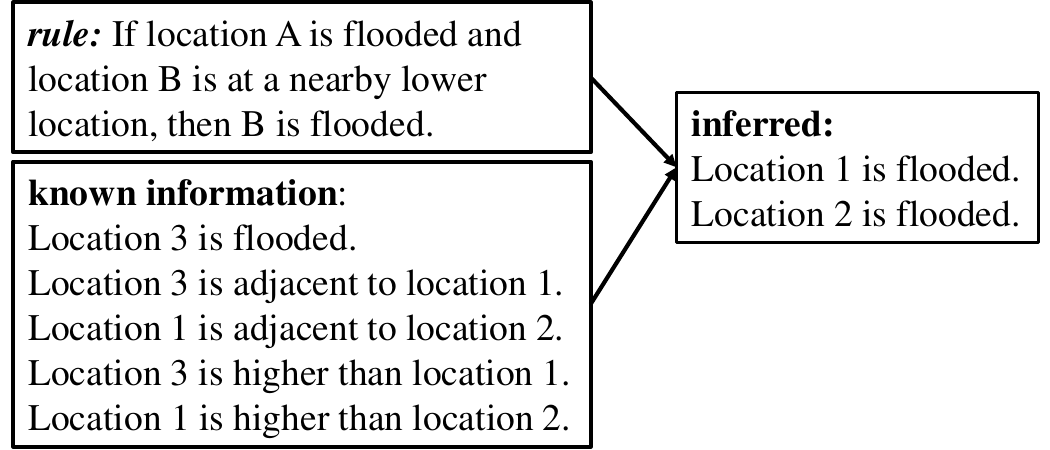}
    \label{fig:problem_infer}
    }
    \caption{Spatial knowledge-infused problem illustration.}
\label{fig:problem}
\end{figure}

Given Earth imagery with limited training labels, a base deep neural network model, and a spatial knowledge base with label constraints, our problem is to infer the full labels while training the neural network. Figure~\ref{fig:problem} provides an example of flooding extent mapping. Data samples are Earth imagery pixels in a raster grid, and explanatory feature layers are spectral bands. Initial noisy labels can be from volunteered geographical information (e.g., geo-tagged tweets). These labels are sparse and limited, as collecting complete high-quality labels through manual annotation is impractical (e.g., high time costs, obscured view due to tree canopies near flood boundary). On the other hand, there exists spatial domain knowledge related to topographical constraints on floodwater distribution, e.g., if location A is flooded and location B is at a nearby lower location, then B is flooded. Similar examples exist in crop type classification \cite{russwurm2017temporal} and tree crown delineation in forest ecology \cite{harmon2022injecting}, land use classification  \cite{jean2019tile2vec}, and ship detection \cite{fabrizio2006knowledge}.

However, the problem poses unique challenges. 
First,  the input labels are spatially sparse and noisy, making it difficult to directly train a neural network on Earth imagery. For example, in flood mapping, in-situ water sensors are often located at a few locations. Second, spatial uncertainty is inherent in the knowledge-guided label inference process, which comes from the noise and sparsity of input labels, imperfect knowledge and rules, and grounding spatial rules on a coarse grid. Third, there are high computational costs associated with spatial logic inference on a large number of raster pixels and a trade-off has to be made between computational efficiency and spatial granularity.
The most closely related works are neural-symbolic systems, which integrate symbolic logical reasoning with deep neural networks \cite{garcez2022neural}. Existing methods focus on replacing the search process of symbolic reasoning to neural network \cite{qu2019probabilistic, zhang2019efficient, marra2021neural},  convert unstructured data, {\em e.g.}, images into symbols for relational learning \cite{yu2022probabilistic}, representing symbolic knowledge as a loss regularization term \cite{hu2016harnessing, diligenti2017semantic, donadello2017logic, xu2018semantic, xie2019embedding, zhou2021clinical, cai2022mitigating}, 
or the combination of logic inference of pseudo-labels and neural network training iteratively \cite{manhaeve2018deepproblog, weber2019nlprolog, dai2019bridging, huang2020semi, li2020closed, tian2022weakly}. However, these methods do not fully address the inherent challenges of spatial data, such as spatial uncertainty and the substantial computational burdens associated with logical inference over a massive number of samples (pixels).

To address the limitations of existing works, we propose \textbf{S}patial \textbf{K}nowledge-\textbf{I}nfused \textbf{H}ierarchical \textbf{L}earning (SKI-HL) that integrates deep learning techniques with spatial knowledge-infused label inference \cite{kimmig2012short, bach2017hinge}. SKI-HL consists of two main modules: the uncertainty-guided hierarchical label inference module and the uncertainty-aware deep learning module. 
The uncertainty-guided hierarchical label inference module captures spatial relationships and dependencies based on a spatial knowledge base and infers labels with quantified uncertainty. To handle the continuous space issue, we design a multi-resolution hierarchy to iteratively refine labels with a trade-off between granularity and computational efficiency. 
The uncertainty-aware deep learning module leverages complete but uncertain labels from the label inference module, capturing information from the data features that cannot be obtained through logical reasoning alone. 
Both modules are trained iteratively to refine inferred labels, reduce uncertainty, and improve deep learning model performance. In summary, the contributions of this paper are as follows:
\begin{itemize}
    \item We propose SKI-HL, a spatial knowledge-infused framework that integrates deep learning and logical reasoning to leverage both explanatory features and spatial knowledge derived from domain logic rules.

    \item Our approach is designed to handle uncertainty in both the original labels and the label inference process, making it more robust and reliable.
    
    \item We propose a strategy to balance the trade-off between spatial accuracy and computational efficiency when discretizing continuous spatial spaces for constructing logic rules and training deep learning models.
    
    \item Taking the flood mapping problem as an example, extensive experiments on real-world datasets demonstrate the superior performance of our model compared to baseline methods.
\end{itemize}

%% file: problem.tex
\section{Problem Statement}\label{sec:prob}
\subsection{Preliminaries}

{\bf Spatial Raster Framework:} 
A spatial raster framework is a tessellation of a two-dimensional plane into a regular grid of $N$ cells. The framework can consist of $m$ non-spatial explanatory \emph{feature layers} and \emph{one class layer}. 
We denote the explanatory feature layers by $\mathbf{X} = \{\mathbf{x}_1, \mathbf{x}_2, \cdots, \mathbf{x}_N \} $ and the class layer by $\mathbf{Y}=\{y_1,y_2,\cdots,y_N\}$, where $\mathbf{x}_i\in\mathbb{R}^{m\times1}$ and $y_i$ are the explanatory features, and class at cell $i$ respectively.  Each cell in a raster framework is a spatial data sample, note as $\mathbf{s}_i = (\mathbf{x}_i, y_i)$, where $i \in \mathbb{N}, 1 \leq i \leq N$.
For example, in the flood mapping problem, the explanatory features are the spectral bands from remote sensing imagery, the target classes are flood and dry categories, and each pixel in the image is a spatial sample. 

{\bf Spatial Knowledge Base:} 
Given a set of spatial samples, which can represent physical objects, locations, events, or any entity relevant to a spatial task, we can describe the relationships and dependencies between these samples using logical statements. This leads us to the concept of a spatial knowledge base. Before defining it formally, we first introduce some fundamental concepts in the logic framework.

\begin{definition}
A \textbf{predicate} is a relation among objects in the domain or attributes of objects (e.g., $Adjacent$), and an \textbf{atom} is a predicate symbol applied to a tuple of terms (e.g., $Adjacent(s_i, s_j)$).
\end{definition}

\begin{definition}
A \textbf{rule} in logic is a clause recursively constructed from atoms using logical connectives and quantifiers. An example would be: $Flood(s_i) \land Adjacent(s_i, s_j) \to Flood(s_j)$.
\end{definition}

\begin{definition}
A \textbf{ground atom} $a$ and a \textbf{ground rule} $r$ are specific variable instantiations of an atom and rule, respectively. A \textbf{grounding} of an atom or rule is a replacement of all of its arguments by constants.
\end{definition}

With these preliminaries, we are now equipped to formally define a spatial knowledge base, $\mathcal{KB}$:

\begin{definition}
A \textbf{spatial knowledge base} $\mathcal{KB}$ is a set of logic rules: $\mathcal{KB} = \{r_1, r_2, \cdots, r_{|\mathcal{KB}|}\}$. Here, each $r_i$ is a rule that represents a spatial relationship, dependency, or constraint between entities in the set of spatial samples $\mathbf{S}$. The quantity $|\mathcal{KB}|$ represents the number of rules in the spatial knowledge base $\mathcal{KB}$.
\end{definition}

Table \ref{tab:rule_example} provides an example of a spatial knowledge base used for a flood mapping on earth imagery problem. Here the variable $s_i, s_j$ stands for a location in the study area or a pixel of earth imagery. It is important to clarify that these rules are probabilistic in nature, reflecting the likelihood of a flood occurrence under certain conditions, rather than providing an absolute certainty. 
In addition, although the rules are represented here in first-order logic, they can be converted into a Markov Logic Network \cite{richardson2006markov} or Probabilistic Soft Logic framework \cite{kimmig2012short, bach2017hinge}, which are statistical relational models that combines probabilistic graphical models and first-order logic.

\begin{table}
\caption{An example of a spatial knowledge base in flood mapping.}
\begin{tabular}{c}
    \hline
    \textbf{Spatial Rules}\\
    \hline
$\forall s_i, s_j \left( Flood(s_i) \land Adjacent(s_i, s_j) \right) \rightarrow Flood(s_j)$ \\
$\forall s_i, s_j \left( River(s_i) \land Distance(s_i, s_j) \leq d \right) \rightarrow Flood(s_j)$ \\
$\forall s_i, s_j \left( Flood(s_i) \land River(s_i) \land Downstream(s_i, s_j) \right) \rightarrow Flood(s_j)$ \\
$\forall s_i \left( LandCover(s_i, \text{Wetlands}) \land HeavyRain(s_i) \right) \rightarrow Flood(s_i)$ \\
$\forall s_i \left( Slope(s_i) > s \right) \rightarrow \neg Flood(s_i)$ \\
$\forall s_i \left( Elevation(s_i) > e \right) \rightarrow \neg Flood(s_i)$\\
    \hline
\end{tabular}
\label{tab:rule_example}
\end{table}

\subsection{Problem Definition}

Formally, we define our problem as follows:

\noindent{\bf Input:}\\
$\bullet$ A large-scale spatial raster framework with spatial samples $\mathbf{S} = \{\mathbf{s}_1, \mathbf{s}_2, \cdots, \mathbf{s}_N \}$\\
$\bullet$ A set of explanatory feature layers $\mathbf{X}$
in $\mathbf{S}$\\
$\bullet$ A limited set of labels $\mathbf{Y_l} = \{y_1, y_2, \cdots, y_l \}$, usually $l \ll N$, each label associated with a sample in $\mathbf{S}$\\
$\bullet$ A spatial knowledge base $\mathcal{KB}$\\
$\bullet$ A base neural network model (e.g., U-Net \cite{ronneberger2015u})\\
{\bf Output:}\\
$\bullet$ Inferred labels $\mathbf{\hat Y}$ with quantified uncertainty $\mathbf{U}$\\
$\bullet$ A deep learning model $DL: \mathbf{X} \rightarrow \mathbf{Y}$\\ 
{\bf Objective:}\\
$\bullet$ Maximize the consistency between inferred labels $\mathbf{\hat Y}$ and $\mathcal{KB}$\\
$\bullet$ Maximize the prediction accuracy of the deep learning model

Specifically, we assume the raster framework $\mathbf{S}$ contains a large number of pixel samples but only with a limited set of labels $\mathbf{Y_l}$. The main objective is to predict the class layer $\mathbf{Y}$ for all spatial samples. To illustrate, consider the case of flood mapping on earth imagery. In this scenario, the set of spatial samples $\mathbf{S}$ corresponds to Earth imagery pixels. The explanatory feature layers $\mathbf{X}$ are the spectral bands. The label set $\mathbf{Y}$ corresponds to the flood status of each pixel ({\em i.e.}, flooded or not). The spatial knowledge contains domain constraints on flood locations (e.g., terrains and topography), which is used to infer flood labels $\mathbf{\hat Y}$. Considering the errors in the inference process and the imperfect logic rules, uncertainty $\mathbf{U}$ naturally exists in the inferred label. 


%% file: approach.tex
\section{The Proposed Approach}

\begin{figure}
    \centering
    \includegraphics[width=0.48\textwidth]{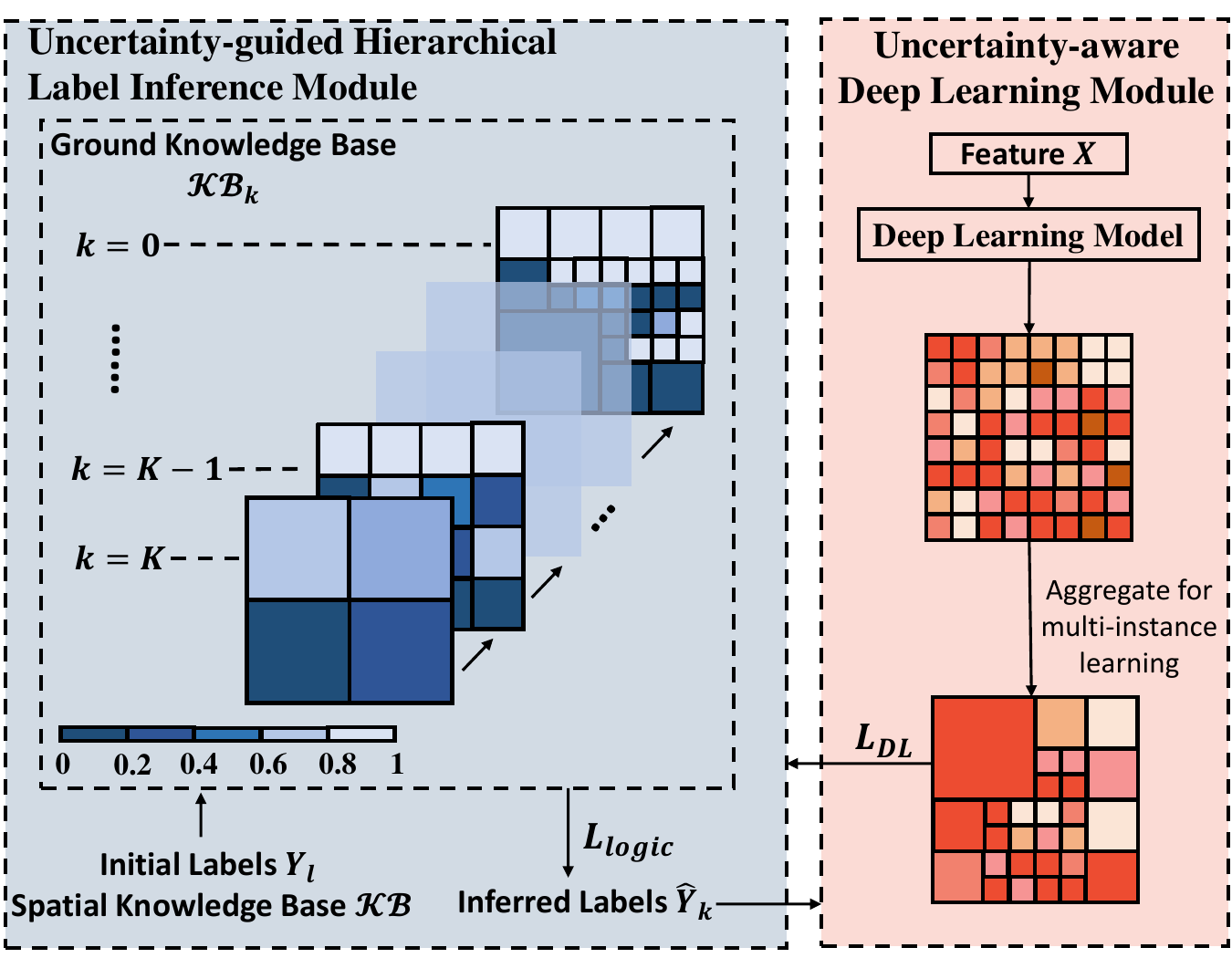}
    \caption{Framework of SKI-HL.}
    \label{fig:framework}
\end{figure}

\subsection{Overview}

Our task is to train the deep learning model and infer sample labels based on the spatial knowledge base. The task is non-trivial for several reasons. First, spatial knowledge inference on labels is computationally expensive due to the immense volume of spatial samples in high-resolution imagery and complex spatial dependencies and interactions. This process quickly becomes computationally infeasible due to the exponential increase in the number of ground atoms and rules. Therefore, scalable grounding strategies are required that can effectively handle these issues by balancing computational efficiency and grounding granularity.  
In addition, the label inference is complicated due to incomplete and sparse initial labels compared with the large study area. Such a low proportion of known data makes logic inference difficult. Furthermore, the labels inferred are not deterministic; instead, they come with uncertainty at different granularity levels, which is non-trivial for the training of a deep learning model.

To address these challenges, we propose a \textbf{S}patial \textbf{K}nowledge-\textbf{I}nfused \textbf{H}ierarchical \textbf{L}earning (SKI-HL) framework. 
Our SKI-HL framework, illustrated in Figure \ref{fig:framework}, consists of two interdependent modules: a hierarchical label inference module and an uncertainty-aware deep learning module. The former infers sample labels in the raster framework with a trade-off between computational efficiency and spatial granularity. We formulate the inference process as an optimization problem with an objective based on the distance loss from Probabilistic Soft Logic (PSL)  \cite{kimmig2012short, bach2017hinge}, a probabilistic logic framework to capture spatial relationships and dependencies, and the spatial grounding configuration in a multi-resolution hierarchical grid structure. We design a greedy heuristic to iteratively refine the inferred labels based on inferred spatial uncertainty. The uncertainty-aware deep learning module trains neural network parameters from uncertain labels in multiple resolutions by an uncertainty-aware loss function and multi-instance learning. The two modules run in iterations: the outputs of the deep learning model will serve as the initialization of the hierarchical label inference module in the next iteration.

\subsection{Hierarchical label inference with spatial knowledge} \label{sec:inference}
The hierarchical label inference module is designed to leverage spatial knowledge within a hierarchical framework to drive the label inference process. Given a raster framework, a spatial knowledge base with logic rules on sample labels, and a subset of initial labels, our goal is to infer the labels of all samples in the framework. 
We need to find a spatial grounding of the rules within the raster framework (e.g., on sample pixels or pixel blocks) and infer the optimal sample class probabilities according to the grounded rules.  For large-scale spatial data (e.g., high-resolution Earth imagery), we need to strike a balance between granularity, computational efficiency, and inference accuracy. If we ground the rules on all high-resolution pixels will lead to too many atoms, making the logic inference computationally expensive. On the other hand, if we ground rules only in low-resolution pixel blocks, the inferred labels are too coarse to train effective deep-learning models. 


We formulate this sub-problem as an optimization problem. There are two specific objectives: the first is to generate accurate sample labels that are consistent with spatial knowledge (expressed by logic rules); the second is to balance computational efficiency and inference granularity in the spatial grounding process.  
We next discuss the specifics of our optimization objective and the proposed greedy strategy.




\subsubsection{Optimization objective}


\begin{figure}[h]
    \centering
    \subfloat[An example for three resolution levels.]{
    \includegraphics[width=0.4\textwidth]{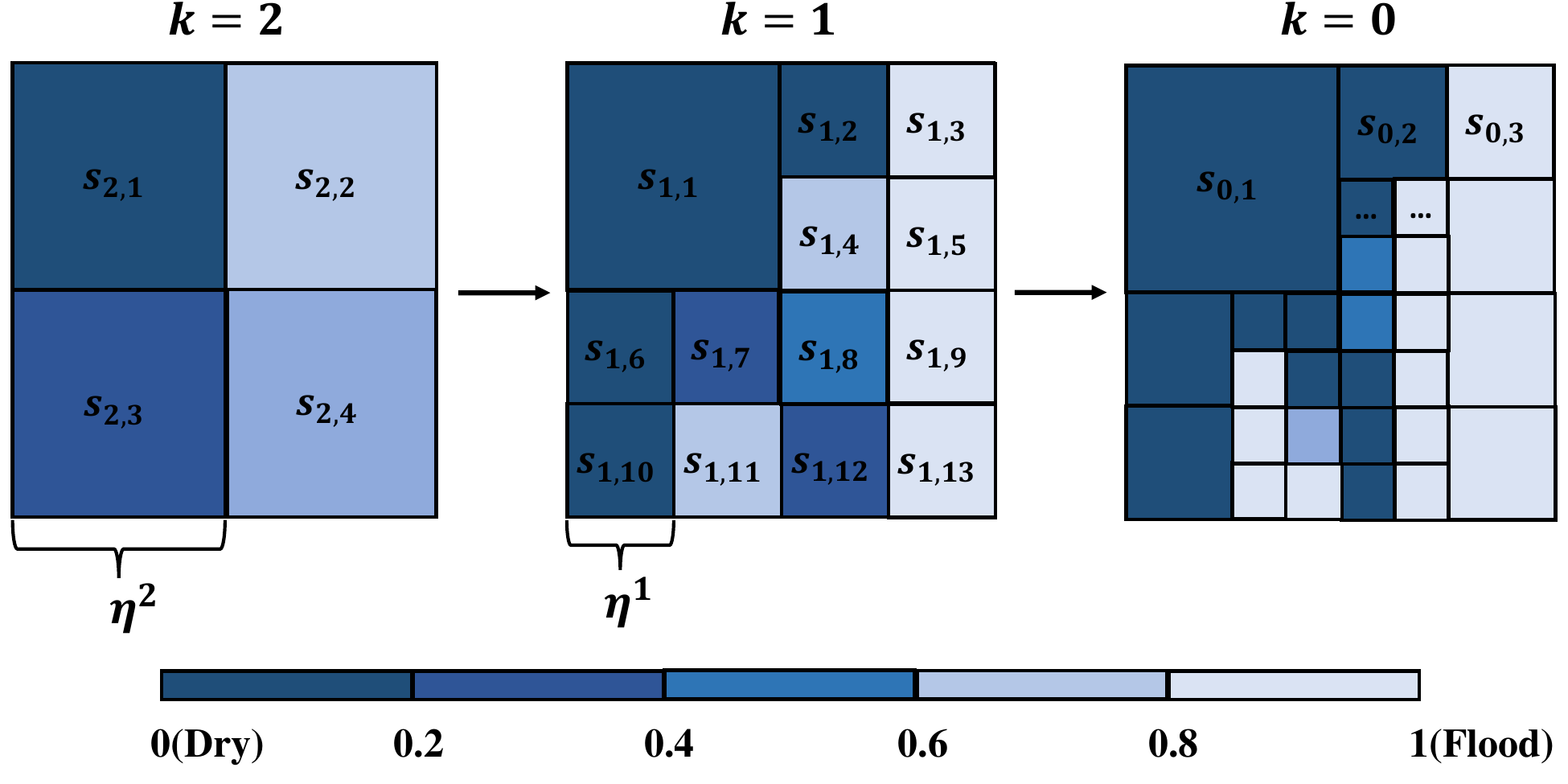}
    \label{fig:three_layer}
    }   
    \\
  \subfloat[Tree structure of the 3-layer hierarchy.]{
    \includegraphics[width=0.4\textwidth]{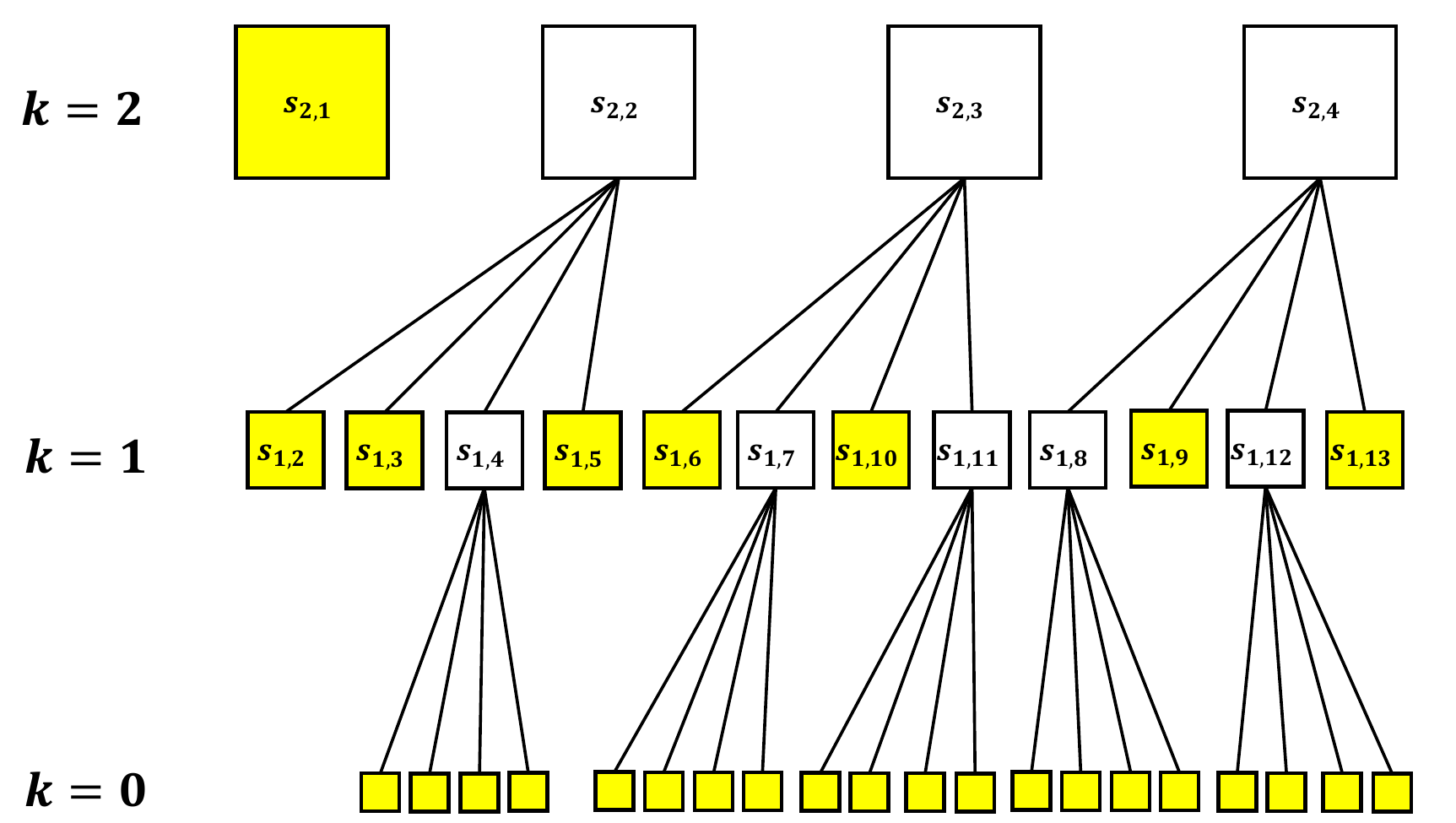}
    \label{fig:tree}
    }
    \caption{Illustration of hierarchical structure.}
\label{fig:hierarchy}
\end{figure}







We now formulate the spatial logic inference of sample labels in a raster framework as an optimization problem. First, we need to define the candidate feasible solution of spatial grounding. The process of spatial grounding refers to substituting the variables in the knowledge base rules with specific, concrete instances, which in our case are spatial samples such as pixels in earth imagery. Given a set of spatial samples and rules from a spatial knowledge base, we substitute each possible sample into the rules to generate candidate feasible solutions for spatial grounding.
Let's illustrate this with an example. Suppose we have a spatial raster comprising 1 million pixels (size of 1000 by 1000) and a spatial logic rule that involves two variables. The process of grounding would mean substituting these variables with every possible pair of pixels. Thus, if we exhaust all possible substitutions, we would end up with a whopping 1 trillion ground rules, derived from each possible combination of the 1 million pixels.

{\bf Spatial hierarchical structure:} Based on the observation that spatial relationships often exhibit hierarchical characteristics and fractal patterns, we exploit a hierarchical framework to address these challenges. Given a large-scale spatial raster framework composed of pixels, we can represent the raster at multiple resolutions. At the coarse level, we can treat each cell as a condensed representation of many pixels. For example, in Figure \ref{fig:three_layer}, if the original raster is $8 \times 8$ pixels (the rightmost grid), a coarser level representation (the leftmost grid) could be a $2 \times 2$ cell grid, where each cell represents a $2 \times 2$ pixel area of the original raster.

In the hierarchical structure of the spatial raster, the coarsest level grid represents the root of the hierarchy, and each subsequent refinement to a finer resolution represents a branching in the hierarchy. As shown in Figure \ref{fig:tree}, in this tree-like structure, each node represents a region with different sizes of the original spatial raster at the finest resolution. With the hierarchical structure, we can start the logic inference from coarse resolution, which decreases the proportion of unlabeled data and makes the inference practical. In the following, we provide a more formal description of the architecture.

Let $\mathbf{S}_0$ be an original large-scale spatial raster framework at the finest resolution. We define a constant $\eta$ where $\eta \in \mathbb N$ and $\eta > 1$, then we can get a set of resolutions $\{1, \eta, \eta^2 \ldots, \eta^K\}$ which denotes the grid sizes at $K + 1$ levels. We use the power number to represent the indexes of the layers $k$ in the hierarchical structure, {\i.e.}, $k = 0, 1, 2, \ldots, K$. The set of samples in each layer $\mathbf{S}_k$, for $k \in \{1, \cdots, K\}$, represents the spatial raster at a specific resolution. The highest layer $K$ corresponds to the coarsest resolution and the lowest layer $0$ corresponds to the finest resolution, usually the original resolution. Each spatial sample $s_{k,i}$ at layer $k$ corresponds to a group of cells at the next finer layer $k-1$. To exemplify the hierarchy, consider Figure \ref{fig:hierarchy} representing 3 varying resolution levels. In this example, the grid size constant $\eta=2$ and we have $3$ layers in the hierarchy, {\em i.e.}, $k=0, 1, 2$. We select a subset of cells from layer 2 to "zoom in" further into smaller cells. This partitioning process results in layer 1, which contains both the original cells and the divided ones. The process is then repeated on the second grid, creating layer 0, which symbolizes the finest resolution layer. Through this hierarchical approach, each cell $s_{k,i}$ in layer $k$ corresponds to a group of cells in the next finer layer $k-1$, {\em e.g.}, cell $s_{2,2}$ corresponds to cell $s_{1,2}, s_{1,3},s_{1,4}$, and $s_{1,5}$ in Figure \ref{fig:hierarchy}. At a certain level, we only need to ground the "leaf node", as shown in Figure \ref{fig:tree}, which significantly decreases the number of ground atoms. Each layer in the hierarchy form a ground knowledge base $\mathcal{KB}_k  = \{r_{k, 1}, r_{k, 2}, \cdots, r_{k, |\mathcal{KB}_k|}\}$ and then generates a set of inferred labels $\mathbf{\hat Y}_k = \{\hat y_{k,1}, \hat y_{k,2}, \cdots, \hat y_{k,N_k}\}$ and their corresponding uncertainties $\mathbf{U}_k = \{u_{k,1}, u_{k,2}, \cdots, u_{k,N_k}\}$, where $N_k$ is the number of samples, {\em i.e.}, grid cells at the $k$-th resolution level.


Second, we need to define the loss function based on the spatial grounding and inferred label probabilities.
To make inferences that are consistent with spatial knowledge, we adopt t-norm fuzzy logic to define the extent of a rule as satisfied, which relaxes binary truth values to a continuous value between $[0, 1]$. These relaxations are exact at the extremes but provide a consistent mapping for values in between. The logical conjunction ($\land$), disjunction ($\vee$) and negation ($\neg$) are as follows:
\begin{equation}
\begin{split}
        I(a_1 \land a_2) & = \max \{I(a_1) + I(a_2) - 1, 0\} \\ 
        I(a_1 \vee a_2) & = \min \{I(a_1) + I(a_2), 1\} \\
        I(\neg a_1) & = 1 - I(a_1)
\end{split}
\label{eq:fuzzy_logic}
\end{equation}
where $I$ is the soft truth value function that can map an atom $a$ or a rule $r$ to an interval between $[0, 1]$, indicating the probability that the atoms or rule holds. Then, following the structure of Probabilistic Soft Logic (PSL), we can induce the distance $d_r(I)$ to satisfaction for a rule $r: r_{body} \rightarrow r_{head}$:
\begin{equation}
    d_r(I) = \max\{0, I(r_{body}) - I(r_{head})\}
\label{eq:distance}
\end{equation}
where $r$ is composed of atoms or negative atoms. PSL determines a rule $r$ as satisfied when the truth value of $I(r_{body}) - I(r_{head}) \geq 0$. To this end, we can convert logical sentences into convex combinations of individual differentiable loss functions, which not only improves the training robustness but also ensures monotonicity with respect to logical entailment, {\em i.e.}, the smaller the loss, the higher the satisfaction. Therefore, given a set of ground rules in the ground knowledge base $\mathcal{KB}$, we can obtain the truth value for all ground atoms, which can serve as inferred labels for the spatial samples.   
\begin{equation}
    \mathbf{\hat Y} = \argmin_{I} \sum_{r \in \mathcal{KB}} \omega_r d_r(I)
\label{eq:infer}
\end{equation}
where $\omega_r$ is the weight of rule $r$. It is noted that here the inferred labels $\mathbf{\hat Y}$ are not binary values but soft truth values between $[0,1]$.

Therefore, in our hierarchical framework, our optimization problem can be summarized as searching for an optimal grounding strategy and minimizing the overall distance to satisfaction for the ground atoms. we formally define the objective to minimize in the hierarchical label inference module as:
\begin{equation}
    L_{logic} = \sum_{k=1}^K \left( \sum_{r \in \mathcal{KB}_k} \omega_r d_r(I_k) + \lambda |\mathcal{KB}_k|) \right)
\end{equation}
where $\mathcal{KB}_k$ stands for the ground knowledge base in the $k$-th layer, $\lambda$ is a balancing coefficient. The summation over $k$ stands for the overall objective of all layers in the hierarchical structure. The first term is the loss defined by PSL distance, which can drive accurate inference. The second term is used to decrease the ground atoms in each layer. 

\subsubsection{A greedy algorithm}
To make a balance between inference accuracy, efficiency, and granularity, we proposed a greedy heuristic grounding strategy. Intuitively, uncertain atom inference always causes a higher distance to the satisfaction of a rule, so here we choose uncertain cells in a coarse layer to refine. The quantified uncertainty $u_{k,i}$ for each cell $i$ at the $k$-th resolution level can be calculated using the entropy of the inferred label $\hat y_{k,i}$ as follows:
\begin{equation}
u_{k,i} = -\hat y_{k,i} \log \hat y_{k,i} - (1 - \hat y_{k,i}) \log (1 - \hat y_{k,i})
\label{eq:uncertainty}
\end{equation}
We select a subset of cells with the highest uncertainty at each resolution level to refine the spatial partitioning. Let $T_k$ be a threshold for selecting high-uncertainty cells at the $k$-th resolution level. We define a set of cells $\{s_{k, i} \mid u_{k,i} \geq T_k\}$ that will be refined to the next finer resolution level $(k-1)$.
Taking layer 2 in Figure \ref{fig:hierarchy} as an example, each cell in this grid is color-coded to denote the probability of dry (dark) and water (light). We view the cell with the leftmost and rightmost color (certain Dry and Flood) in the color bar as certain cells, and others as uncertain cells, {\em i.e.}, only the $s_{2, 1}$ can be view as a certain cell in layer 2.

For the selected cells in $\mathbf{S}_k$, we construct a new spatial partitioning with smaller cell size and update the grounding atoms set accordingly. As shown in Figure \ref{fig:hierarchy}, the uncertain coarser cells in layer 2 are split into $2 \times 2$ finer cells, respectively. We then perform PSL inference using the hierarchical label inference module at the $(k-1)$-th resolution level with only the leaf node in the hierarchy (yellow cells). Since the distance-based loss is convex, we can use gradient descent to optimize it. To initialize $I$ at different resolutions, in the first iteration, {\em i.e.}, we pre-train the deep learning model with limited labels and use the output probabilities as the initialization. In the following iterations, the predicted probabilities of the corresponding deep learning model are regarded as the initial soft truth value of the ground atom in each $r$. 

Starting from the coarsest resolution ($k=K$), the process continues iteratively until the finest resolution ($k=0$) is reached. By employing this iterative hierarchical approach, our framework effectively improves the resolution of sparse labels from coarse to fine while efficiently handling the computational challenges associated with high-resolution spatial analysis and decision-making problems. Moreover, the method focuses computational resources on high-uncertainty regions, ensuring accurate label inference in these critical areas. Then, these inferred labels will serve as the training label in the uncertainty-aware deep learning module, which we will introduce in the next section. The overall process is described in Algorithm \ref{algo:overall}.

\begin{algorithm}[!t]
\caption{SKI-HL algorithm}
\label{algo:overall}
\begin{algorithmic}[1]
 \REQUIRE 

 Spatial knowledge base $\mathcal{KB}$, Maximum resolution level $K$, Resolution constant $\eta$, Existing sparse label $\mathbf{Y_l}$, rule weights $\mathbf{\omega}$, Deep learning model $DL$, Explanatory features $\mathbf{X}$\\

 \ENSURE Fine resolution predicted labels $\hat Y$, trained deep learning model $DL$ \\
 
 \STATE Initialize resolution level $k \leftarrow K$

 \WHILE{$k!=0$}
 \STATE Grounding rules in current resolution level
 \STATE Infer the uncertain labels $\mathbf{\hat Y}_k$ based on Equation \ref{eq:infer}
 \STATE Calculate the uncertainty $\mathbf{U}_k$ based on Equation \ref{eq:uncertainty}
 \STATE Train the deep learning model $DL$ using $\mathbf{\hat Y}_k$ based on Equation \ref{eq:dl_final}
 \STATE $k \leftarrow k - 1$
 \ENDWHILE
 \RETURN $DL$, $\mathbf{\hat Y}_k$
\end{algorithmic}
\end{algorithm}

By observing the hierarchical structure and the figures, it becomes evident that this approach provides a computationally efficient way to handle large-scale spatial data, making a trade-off between granularity and efficiency. This is especially crucial in real-world applications like flood mapping, where both precision and computational resources are of prime importance.

As we mentioned before, there is a fundamental trade-off between computational efficiency and inference granularity in the grounding process. Therefore, given this hierarchical structure, the problem is how to select cells at each level of the hierarchy for refinement to the next finer resolution. We need to select cells for refinement in such a way that balances the need for detailed information (i.e., higher resolution) against the computational cost of refining cells (i.e., efficiency). This problem also involves determining which cells to refine based on the potential benefit in terms of inference accuracy. The challenge lies in making these decisions efficiently and effectively, given the large scale of the spatial raster and the complexity of the spatial relationships represented in the spatial knowledge base.

\subsection{Uncertainty-aware deep learning}
The uncertainty-aware deep learning module, another component of SKI-HL, is capable of capturing information from the explanatory features, which is not accessible via logic inference alone. In addition, it plays a significant role in handling the uncertainty of inferred labels and variations in resolution. In traditional deep learning models, the model makes a prediction for each sample, but it doesn't utilize any information about how confident the model is about that prediction \cite{he2023survey}. This could lead to overconfident predictions in regions with scarce or noisy labels. The module employs a modified version of the Binary Cross Entropy loss, offering a method to manage the uncertainty from the spatial knowledge and inferred label. In this module, all the deep learning predictions are at the finest resolution, {\em i.e.}, under the original spatial raster framework and using original explanatory features.

\subsubsection{Incorporating Uncertainty in Labels}
The initial phase of training employs the Binary Cross Entropy loss function, defined as:
\begin{equation}
L_{DL} = -\sum_{i=1}^{N} y_{i}\log p_{i} + (1-y_{i})\log(1-p_{i}), y_i = 0 \text{ or } 1
\end{equation}
where $y_{i}$ is the ground truth label, $p_{i}$ signifies the predicted probability output by the deep learning model, and $N$ represents the number of spatial samples in a particular layer of the hierarchy.

The key challenge in our problem scenario is dealing with the inherent uncertainty associated with inferred labels. This stems from the fact that we start with a sparse and limited set of labels and then attempt to extrapolate this information to the entire spatial domain. A common approach would be to predict binary labels (flooded or dry) for each location. However, this ignores the degree of confidence (or uncertainty) in these predictions.

To better handle this, we replace the ground truth labels $y_{i}$ in the Binary Cross Entropy loss function with the inferred uncertain labels $\hat{y}_{i}$. This allows us to directly optimize for the accuracy of the predicted probabilities, rather than merely the binary labels. The loss function thus becomes:
\begin{equation}
L_{DL} = -\sum_{i=1}^{N} \hat{y}_{i}\log p_{i} + (1-\hat{y}_{i})\log(1-p_{i})
\end{equation}
where $\hat{y}_{i} \in \mathbf{\hat Y}$ is the inferred uncertain label. Currently, the cross entropy measures the difference between the predicted probability distribution of the deep learning model and the probability distribution of inferred labels. This modification effectively incorporates uncertainty information into the training process and can improve the model's performance when dealing with ambiguous cases. 

\subsubsection{Addressing Multi-instance Learning Scenarios}
In scenarios where we partition the spatial domain into non-overlapping cells, or instances, and construct a hierarchical structure for label inference, we encounter a multi-instance learning scenario \cite{foulds2010review}. Since we have different resolution levels, here, instead of assigning a single label to each pixel, we need to compute an aggregate probability output for each pixel to capture the overall likelihood of the event (e.g., flooding) occurring within the corresponding coarser cell. As shown in the right-bottom of Figure \ref{fig:framework}, to make an alignment between the structure of deep learning output and the inferred label, we have to combine the predictions of some pixels together, {\em e.g,}, aggregate the $4 \times 4$ pixels in the output mask to train with the only 1 label inferred by logical reasoning.

In this context, the loss function should be updated again to reflect the cell-level nature of the labels:
\begin{equation}
L_{DL} = -\sum_{i=1}^{N_k} \hat{y}_{k,i}\log P_{k,i} + (1-\hat{y}_{k,i})\log(1-P_{k,i})
\label{eq:dl_final}
\end{equation}
The probability output $P_{k,i}$ for each cell sample $s_{k,i}$ is computed as:
\begin{equation}
P_{k,i} = \frac{1}{|s_{k,i}|}\sum_{s_{0, j}\in s_{k,i}} p_j
\end{equation}
where $P_{k,i}$ is the average of the predicted probabilities $p_j$ for all the samples $s_j$ within the coarse cell $s_{k,i}$. $|s_{k,i}|$ represents the number of finest resolution pixels in the cell. This is a natural way to aggregate the predictions of individual locations within a cell to produce a cell-level prediction. It allows the model to handle different levels of granularity in the spatial domain, making it flexible and adaptable to various spatial scales.

%% file: evaluation.tex
\section{Evaluation} \label{sec:eval}

\subsection{Experiments setup}

{ \bf Dataset Description:} We use two real-world flood mapping datasets collected from North Carolina during Hurricane Matthew in 2016. The explanatory features comprise the red, green, and blue bands within the aerial imagery obtained from the National Oceanic and Atmospheric Administration's National Geodetic Survey\footnote{https://www.ngs.noaa.gov/}. In addition, digital elevation imagery was sourced from the University of North Carolina Libraries\footnote{https://www.lib.ncsu.edu/gis/elevation}. Each piece of data was subsequently resampled to a 2-meter by 2-meter resolution to standardize the information.
For Dataset 1, the image has a shape of $2500 \times 1800$ with $4.5$ million pixels. For Dataset 2, the image has a shape of $3400 \times 8400$ with $28.56$ million pixels. 
In alignment with the principles of transductive learning, the experiment leverages both explanatory features and spatial information across the entire area throughout the learning process. A sparse set of labeled pixels forms the training set while the test set includes the whole area, excluding the labeled pixels. 

{ \bf Candidate Methods:} In our experiments, we compare our proposed SKI-HL model with a variety of baselines that represent different approaches to handling spatial data and infusing knowledge into deep learning. 

\begin{itemize}
    \item \textbf{Pretrain:} In this method, the deep learning model is trained with the initially labeled pixels for each dataset. 
    \item \textbf{Self-training:} The model adds patches with high confidence from Pretrain to the training dataset and re-trains the model. 
    \item \textbf{DeepProbLog \cite{manhaeve2018deepproblog}:} This is a programming language that integrates deep learning with probabilistic logic programming. It allows for the incorporation of neural networks within a logic program, and these neural networks can be used to define probabilistic facts. 
    \item \textbf{Abductive Learning (ABL) \cite{dai2019bridging}:} This is a learning framework that combines both reasoning and learning. It works by training a model to make predictions and then using a logic reasoner to validate these predictions against a set of given logic rules. If the prediction contradicts the rules, the learning algorithm will revise its model based on the abductive explanation
    \item  \textbf{SKI-HL-Base} This is a simplified version of our model as a candidate method that doesn't implement the selection of uncertain areas in the grounding process. Instead, it ground all atoms for each layer in the hierarchy. 
\end{itemize}

\begin{table}\footnotesize
\centering
\caption{Accuracy versus Uncertainty (AvU).}
\begin{tabular}{|c|c|c|c|}
\hline
 &  &\multicolumn{2}{|c|}{Uncertainty}  \\ \hline
 & & Certain&Uncertain  
 \\ \hline{}

\multirow{2}{*}{Accuracy} &Accurate & Accurate Certain (AC)&{ Accurate Uncertain  (AU)}\\ \cline{2-4} 
&{ Inaccurate}&{Inaccurate Certain (IC)}&{Inaccurate Uncertain (IU)}\\ \hline

\end{tabular}
\label{tab:avutable}
\end{table}

\textbf{Classification evaluation metrics:} We used precision, recall, and F1 score on the flood mapping class to evaluate the pixel-level classification performance.

\textbf{Uncertainty quantification evaluation metrics:} The performance of uncertainty estimations in our model is quantitatively evaluated using the Accuracy versus Uncertainty ($AvU$) measure, as shown in previous work \cite{krishnan2020improving, he2022quantifying}. We set an uncertainty threshold, denoted by $T_{u}$, to group uncertainty estimations into 'certain' and 'uncertain' categories. Predictions based on these estimations are then grouped into four categories: Accurate-Certain ($\text{AC}$), Accurate-Uncertain ($\text{AU}$), Inaccurate-Certain ($\text{IC}$), and Inaccurate-Uncertain ($\text{IU}$).
Let $n_{\text{AC}}, n_{\text{AU}},n_{\text{IC}}, n_{\text{IU}}$ represent the number of samples in the respective categories. The $AvU$ measure evaluates the proportion of $\text{AC}$ and $\text{IU}$ samples, with the idea being that accurate predictions should ideally be accompanied by certainty, and inaccurate predictions should correspondingly indicate uncertainty. This measure lies in the range $[0, 1]$, with higher values indicating more reliable model performance.
Specifically, we compute $AvU_{A}$ for accurate predictions and $AvU_{I}$ for inaccurate predictions as follows:
\begin{equation}\footnotesize
\begin{split}
AvU_{A} = \frac{n_{\text{AC}} }{n_{\text{AC}} + n_{\text{AU}}},
AvU_{I} = \frac{n_{\text{IU}}}{ n_{\text{IC}} + n_{\text{IU}} }
\end{split}
\end{equation}
In our evaluation, we compute the harmonic average of $AvU_{A}$ and $AvU_{I}$ to penalize extreme cases:
\begin{equation}\label{eq:avu}\footnotesize
AvU = \frac{2*AvU_{A}*AvU_{I}}{AvU_{A} + AvU_{I}}
\end{equation}

This evaluation approach thus offers a comprehensive measure of the reliability of our model's uncertainty estimations.

{\bf Model configuration:} When implementing our method and baselines, we considered U-Net, a powerful deep learning model for image segmentation, as the base model. We set the same set of architecture for the U-Net model in all baselines with 5 downsample operations and 5 upsample operations. There is a batch normalization within each convolutional layer and the dropout rate is 0.2.

For Pretrain, images are divided into 100 by 100 patches, using patches containing a labeled pixel for training. All pixels in these patches are assigned the label of the initially labeled pixel for pre-training. Self-training uses Pretrain predictions to enhance the training dataset, iteratively adding high-confidence patches based on average predicted class probabilities. 
The same pretrained U-Net initializes the deep learning models in DeepProbLog, ABL, and our proposed SKI-HL frameworks. For DeepProbLog, the time cost of pixel-level inference is intractable, so we can only conduct patch-level inference.

For the hierarchical structure of SKI-HL, we set the grid size constant $\eta = 10$ and $K=2$ which means there are 3 layers with grid size $100$, $10$, and $1$ respectively in this hierarchy. We construct the spatial knowledge base for the flood mapping task based on distance and topology relationships. For the distance relationship, we directly use the neighborhood pair to model. For the elevation, we adopt a Hidden Markov Tree model \cite{xie2018geographical, jiang2019geographical} which can model the topological relationship of each location based on the elevation. 

\subsection{Comparison on classification performance}

\begin{table*}
\caption{Comparison on classification.}
\begin{tabular}{|l|c|ccccc|ccccc|}
\hline
\multirow{2}*{Method} &  \multirow{2}*{Class} & \multicolumn{5}{c|}{Dataset 1} & \multicolumn{5}{c|}{Dataset 2} \\ \cline{3-12}
    &  & P & R & F1 & Avg. F1 & Acc & P & R & F1 & Avg. F1 & Acc \\
\hline
 
\multirow{2}{*}{Pretrain}& Dry & 0.79 & 0.62 & 0.70 & \multirow{2}{*}{0.74} & \multirow{2}{*}{0.75}  & 0.88 & 0.76 & 0.81 & \multirow{2}{*}{0.74} & \multirow{2}{*}{0.76} \\ 
                                   & Flood & 0.73 & 0.86 & 0.79 & & & 0.59 & 0.77 & 0.67 & &\\ 
\hline
\multirow{2}{*}{Self-training}& Dry & 0.60 & 0.83 & 0.70 & \multirow{2}{*}{0.78} & \multirow{2}{*}{0.81} & 0.89 & 0.77 & 0.83 & \multirow{2}{*}{0.76} & \multirow{2}{*}{0.78} \\ 
                                   & Flood & 0.93 & 0.81 & 0.86 & & & 0.61 & 0.80 & 0.69 & &\\         
\hline
\multirow{2}{*}{DeepProbLog}& Dry & 0.73 & 0.78 & 0.75 & \multirow{2}{*}{0.81} & \multirow{2}{*}{0.83} & 0.83 & 0.87 & 0.85 & \multirow{2}{*}{0.82} & \multirow{2}{*}{0.82}\\ 
                                   & Flood & 0.88 & 0.85 & 0.87 &  & & 0.82 & 0.77 & 0.79 & &\\     
                                   
\hline
\multirow{2}{*}{ABL}& Dry & 0.66 & 0.78 & 0.72 & \multirow{2}{*}{0.79} & \multirow{2}{*}{0.81} & 0.76 & 0.86 & 0.81 & \multirow{2}{*}{0.79} & \multirow{2}{*}{0.79} \\ 
                                   & Flood & 0.90 & 0.83 & 0.86 & & & 0.82 & 0.71 & 0.76 & &\\   
\hline

\multirow{2}{*}{SKI-HL-Base}& Dry & 0.95 & 0.93 & 0.94 & \multirow{2}{*}{\bf 0.95} & \multirow{2}{*}{\bf 0.95} & 0.92 & 0.94 & 0.93 & \multirow{2}{*}{0.91} & \multirow{2}{*}{0.91} \\ 
                                   & Flood & 0.96 & 0.97 & 0.96 & & & 0.91 & 0.88 & 0.90 & &\\                                    
\hline
\multirow{2}{*}{SKI-HL}& Dry & 0.96 & 0.92 & 0.94 & \multirow{2}{*}{\bf 0.95} & \multirow{2}{*}{\bf 0.95} & 0.91 & 0.97 & 0.94 & \multirow{2}{*}{\bf 0.92} & \multirow{2}{*}{\bf 0.93}\\ 
                                   & Flood  & 0.95 & 0.98 & 0.96 & & & 0.95 & 0.88 & 0.91 &  &\\ 
\hline
     
\end{tabular}
\label{tab:performance_1}
\end{table*}

We first test the performance of each model with 4 labeled pixels. The experimental results shown in Table~\ref{tab:performance_1} underscore the efficacy of our proposed SKI-HL approach over the baseline models. The Pretrain model shows the poorest performance with a small labeled dataset, owing to the numerous surface obstacles confusing the classifier, limiting its generalization capabilities for large-scale data. While the Self-training model outperforms the Pretrain model, its high-confidence predictions may still contain errors, and its inability to integrate spatial knowledge limits its performance.
We notice that both DeepProbLog and ABL display better performance, highlighting the importance of integrating spatial knowledge into learning. Furthermore, the performance of ABL is generally lower than DeepProbLog, the gap is due to that ABL only use first-order logic as hard constraints to revise label and cannot deal with complex spatial rules which inherently have uncertainty and cannot always be perfectly satisfied. Nevertheless, DeepProbLog, not originally designed for large-scale spatial problems, necessitates patch-level inference due to time complexity, diminishing its performance.

Our proposed SKI-HL method and its base model consistently outperform the baseline models on both datasets. Despite not grounding all pixels in the SKI-HL model, it achieves parallel performance on Dataset 1 with its base model and even surpasses it on Dataset 2. The performance enhancement is due to the uncertainty-guided hierarchical label inference, which enriches the model's capacity to encapsulate complex spatial dependencies and dynamically concentrate on areas of uncertainty for pixel-level inference. Further, the hierarchical structure eases the demand for complete dense labeling, a bottleneck for other baseline models.
Taken together, the superior performance of the SKI-HL approach accentuates the importance of efficiently integrating spatial domain knowledge with deep learning and harnessing the benefits of hierarchical label inference, especially in the context of large-scale spatial applications with limited training labels.

\subsection{The effect of the number of initial labeled samples}
\begin{figure}
    \centering
{
    \includegraphics[width=0.4\textwidth]{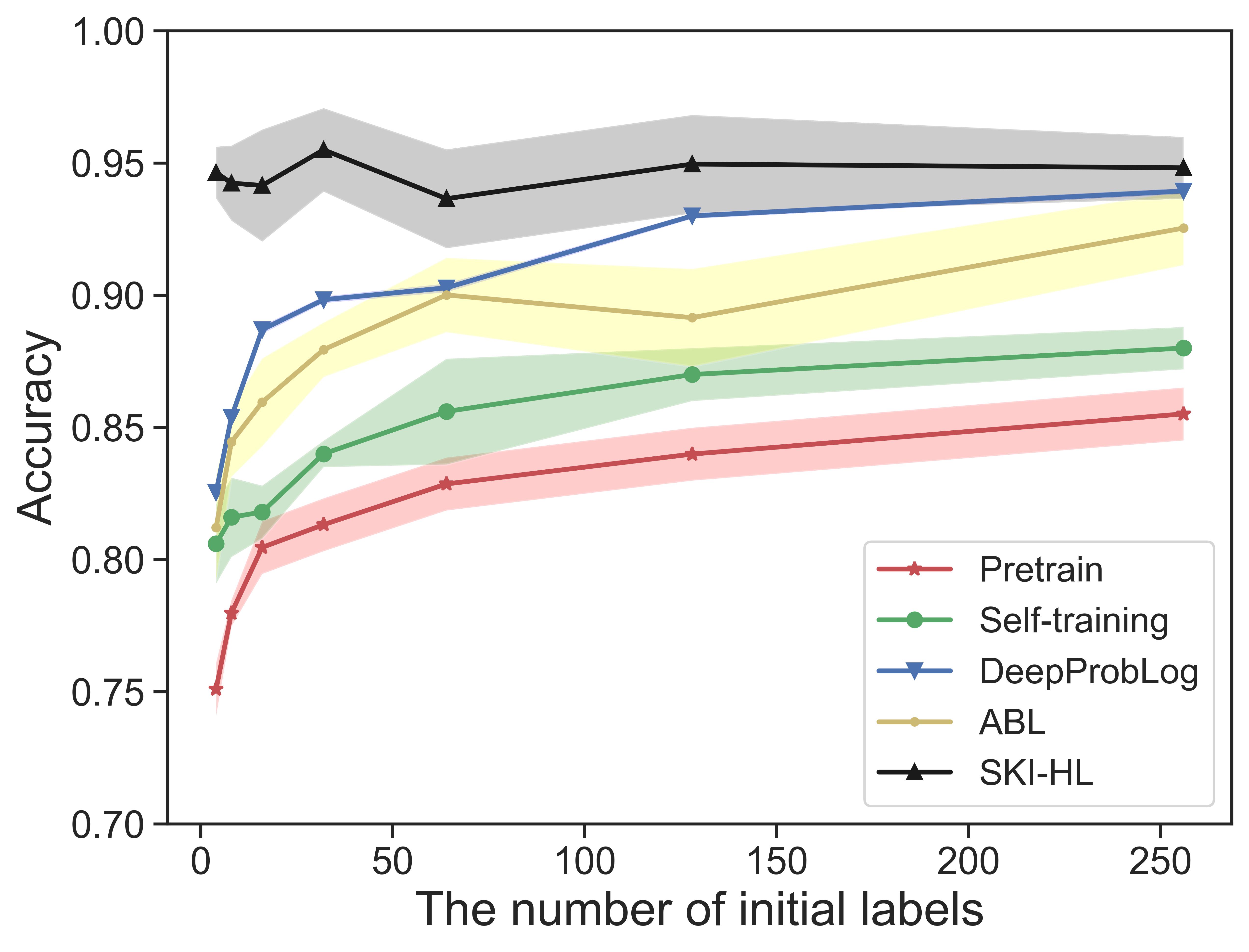}
    }   
    \caption{Accuracy comparison on different numbers of initial labels.}
\label{fig:seed_effect}
\end{figure}

In order to thoroughly evaluate the effectiveness of our proposed SKI-HL method, we conduct experiments using Dataset 1 with varying numbers of initial labeled data, ranging from 4 to 256 in multiples of 2. The classification accuracy over 5 runs is utilized as the evaluation metric, with the mean value represented by markers and standard deviation indicated by shadows in Figure \ref{fig:seed_effect}. 
All baseline models display an upward trend in accuracy with the increase in initial labels. However, the Pretrain and Self-training models exhibit less significant improvement. This phenomenon can be attributed to the fact that these models use labeled pixels to represent the corresponding whole patches during the training process. 
While ABL's performance does show improvement with an increased number of labels, its limited ability to handle complex spatial rules due to the usage of hard logic for label revision makes it unable to reconcile potential conflicts, and then fail to achieve higher performance.
DeepProbLog's accuracy incrementally improves and eventually stabilizes at a higher level as the number of initial labels increases. Owing to its logic-based framework for computing gradients, it consistently yields stable results with negligible variance.
In contrast to the other models, our SKI-HL framework remains relatively unaffected by the number of initial labels, maintaining an accuracy rate of around 0.95. The underlying reason lies in our model's unique label inference process: it starts from a coarse resolution and subsequently refines labels, ensuring accurate label inference. The slight variations in the results can be attributed to training randomness and the initialization process. 

\subsection{Comparison on uncertainty quantification performance}

\begin{table}[ht]\footnotesize
\caption{Comparison on uncertainty quantification.}
\begin{tabular}{|l|c|cc|cc|}
\hline
\multirow{2}*{Method} &  \multirow{2}*{Accuracy} & \multicolumn{2}{|c|}{Dataset 1} & \multicolumn{2}{|c|}{Dataset 2} \\  \cline{3-6}
    &  & $AvU_{A}/AvU_{I}$ & $AvU$ & $AvU_{A}/AvU_{I}$ & $AvU$ \\
\hline
 
\multirow{2}{*}{Pretrain}& Accurate & 0.81 & \multirow{2}{*}{0.45} & 0.97 & \multirow{2}{*}{0.20}\\ 
                                   & Inaccurate & 0.32 &  & 0.11 & \\ 
\hline
\multirow{2}{*}{Self-training}& Accurate & 0.85 & \multirow{2}{*}{0.65}& 0.92 & \multirow{2}{*}{0.37}\\ 
                                   & Inaccurate &  0.53 & & 0.23 &  \\ 
\hline
\multirow{2}{*}{DeepProbLog}& Accurate & 0.90 & \multirow{2}{*}{0.40} & 0.79 & \multirow{2}{*}{0.61}\\ 
                                   & Inaccurate & 0.26 & & 0.50 &   \\              
\hline
                            
\multirow{2}{*}{ABL}& Accurate & 0.85 & \multirow{2}{*}{0.44} & 0.91 & \multirow{2}{*}{0.43}\\ 
                                   & Inaccurate & 0.29 & & 0.28 &   \\  

\hline
\multirow{2}{*}{SKI-HL-Base}& Accurate & 0.82 & \multirow{2}{*}{0.69} & 0.51 & \multirow{2}{*}{0.64}\\ 
                                   & Inaccurate & 0.59 & & 0.86 &\\                                    
\hline
\multirow{2}{*}{SKI-HL}& Accurate & 0.80 & \multirow{2}{*}{\bf 0.74} & 0.80 & \multirow{2}{*}{\bf 0.84}\\ 
                                   & Inaccurate & 0.68 &  & 0.88 & \\ 
\hline
     
\end{tabular}
\label{tab:uncertainty_1}
\end{table}

In Table \ref{tab:uncertainty_1}, we notice a clear distinction in uncertainty estimation between our proposed SKI-HL model and the baselines. While Pretrain and Self-training models manifest a larger gap between $AvU_{A}$ and $AvU_{I}$, this discrepancy is mitigated in DeepProbLog and ABL, which effectively incorporate spatial knowledge into learning. However, they still struggle to achieve a balanced $AvU_{A}$ and $AvU_{I}$, particularly in situations of sparse and noisy labels. In stark contrast, our proposed SKI-HL model exhibits a superior performance on both datasets, signifying its robust ability to model complex spatial dependencies and adjust to areas of uncertainty dynamically. The integration of uncertainty-guided hierarchical label inference further mitigates the impact of sparse labeling, a bottleneck for other models. This finding emphasizes the pivotal role of efficiently integrating spatial domain knowledge with deep learning, especially under the constraints of limited training labels, in achieving reliable uncertainty estimation for large-scale spatial applications.

\subsection{Case Study}

\begin{figure}
    \centering
    \subfloat[Image, Ground Truth and Elevation Map]{
    \includegraphics[width=0.4\textwidth]{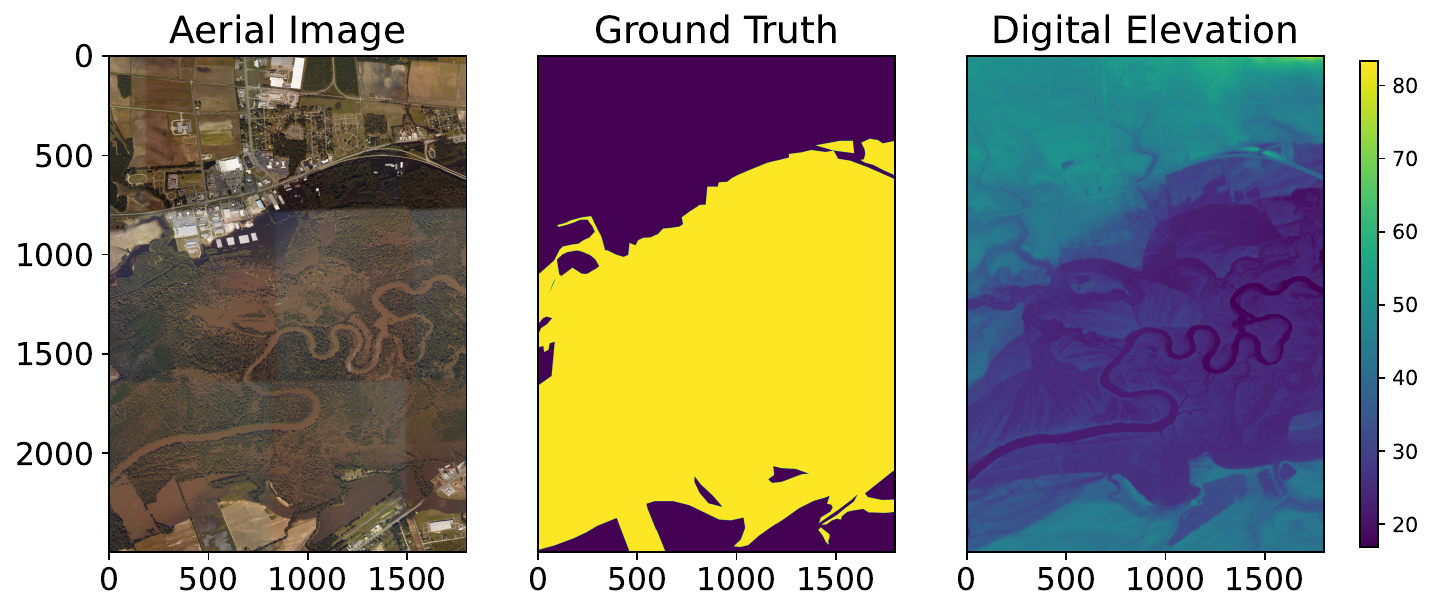}
    \label{fig:visual_1}
    }   
    \\
    \subfloat[Inferred label at different resolution level.]{
    \includegraphics[width=0.4\textwidth]{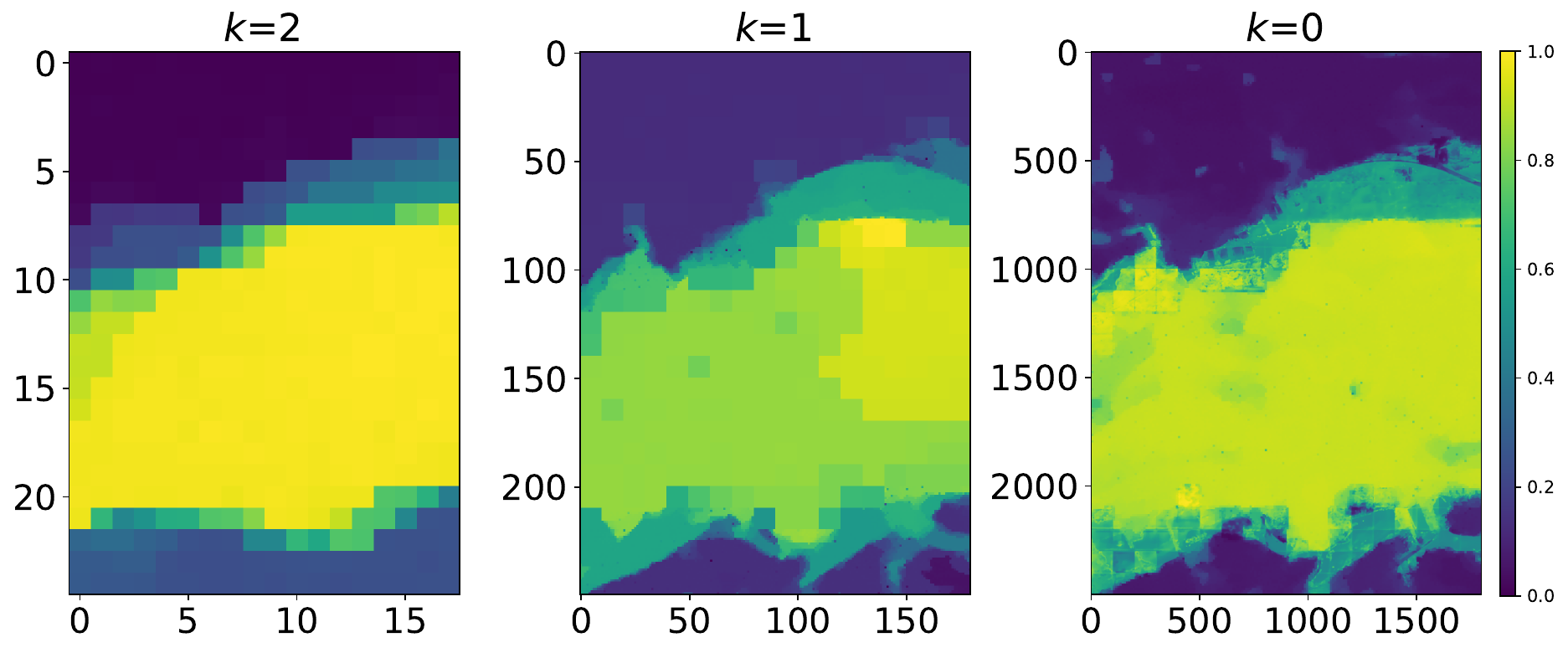}
    \label{fig:visual_2}
    }   
    \\
    \subfloat[Deep learning prediction at different resolution level.]{
    \includegraphics[width=0.4\textwidth]{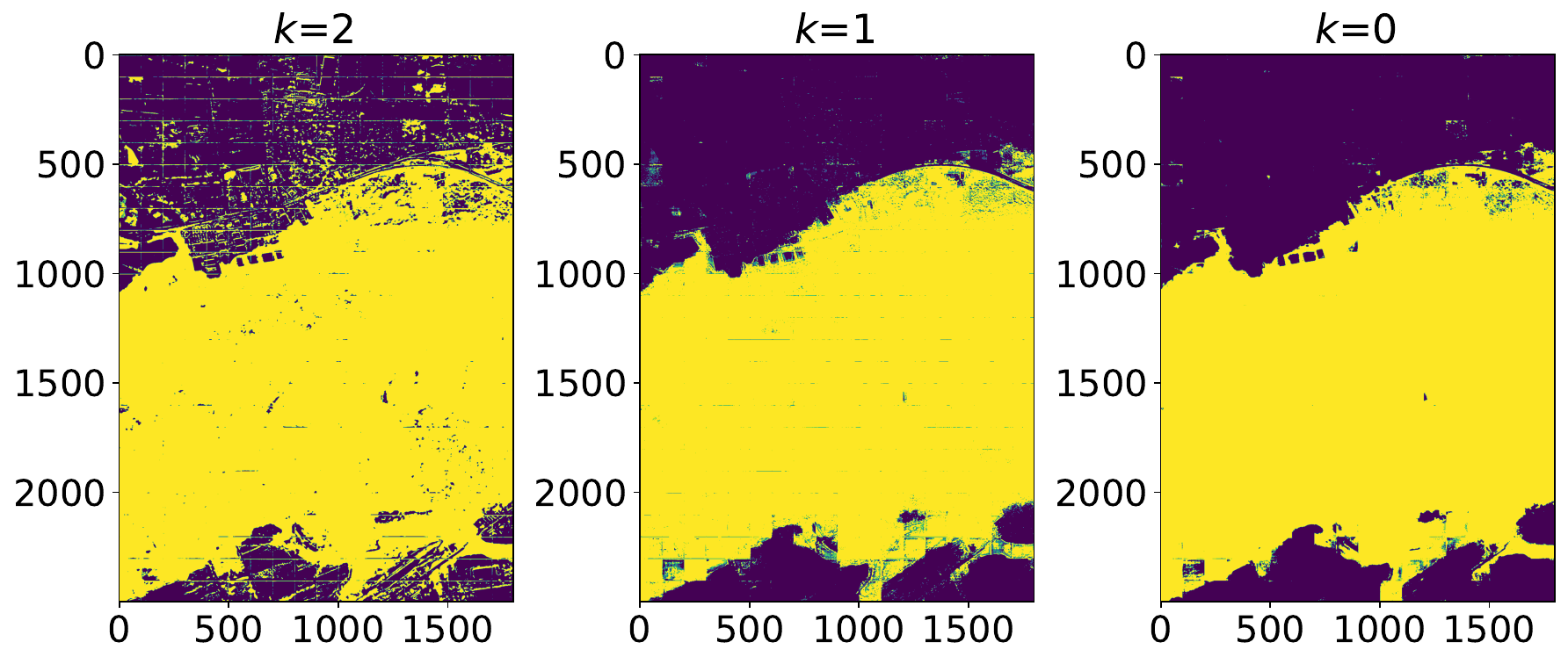}
    \label{fig:visual_3}
    }
    \caption{Performance with different resolution in Dataset 1.}
\label{fig:visual}
\end{figure}

In our case study, we visually analyze the effectiveness of our model across varying resolution levels. As depicted in Figure \ref{fig:visual_1}, we present the aerial earth imagery, ground truth label, and digital elevation map from Dataset 1. It is noted that we don't use the ground truth to train our label, instead, it was only used for testing. Figures \ref{fig:visual_2} and \ref{fig:visual_3} illustrate the evolution of inferred labels and deep learning predictions at different resolution levels.
The resolution of the inferred labels refines progressively from a coarse resolution of 25 by 18 to the finest resolution of 2500 by 1800. This process allows for the accurate detection and refinement of uncertain areas, which often represent flood boundaries.
Simultaneously, the granularity increase of the training labels results in an improved output from the deep learning model. A clear reduction in misclassified pixels can be observed, appearing as noise within each class of the area. This improvement can be attributed to the fact that multi-instance learning, used with coarse resolution labels, cannot provide supervision to every pixel. Hence, as our approach refines the label resolution, the deep learning model is able to generate more accurate predictions.

\subsection{Analysis of time costs with hierarchical label inference}
\begin{figure}
    \centering
    \subfloat[Dataset 1]{
    \includegraphics[width=0.210\textwidth]{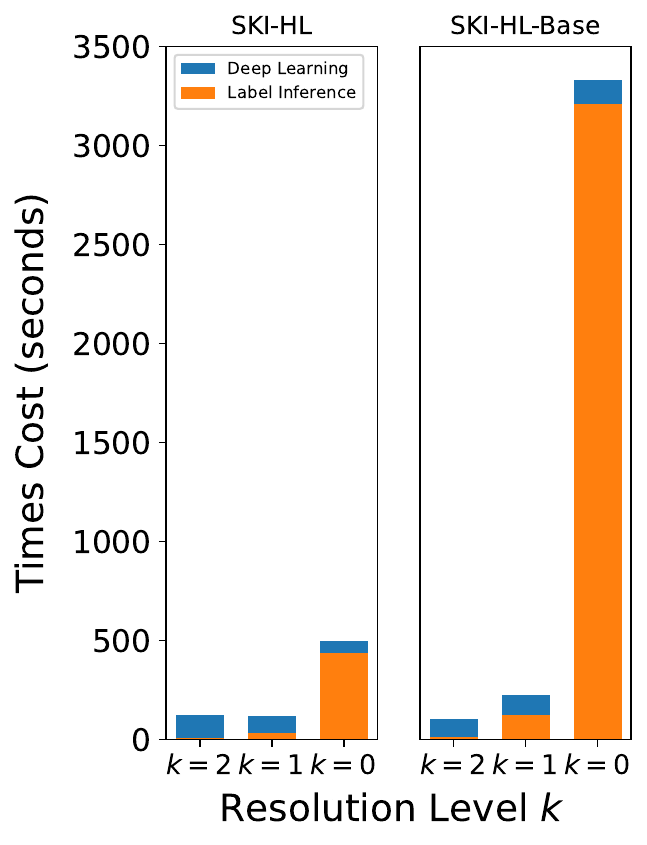}
    }   
    \subfloat[Dataset 2]{
    \includegraphics[width=0.215\textwidth]{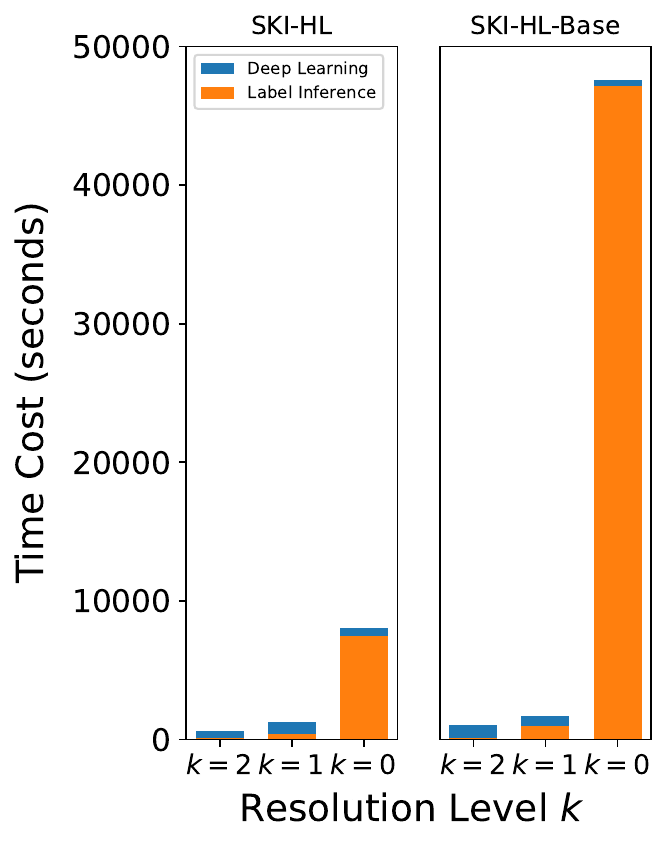}
    }   
    \caption{Comparison on time cost.}
\label{fig:time_cost}
\end{figure}


In order to demonstrate the computational efficiency of our proposed approach, we conducted a set of experiments evaluating the time costs of SKI-HL and its base model across different resolution levels. These experiments were executed on an AMD EPYC 7742 64-Core Processor CPU and an NVIDIA A100 GPU equipped with 80 GB of memory.

Figure \ref{fig:time_cost} presents the time costs associated with the uncertainty-aware deep learning model training (blue bar) and the hierarchical label inference module training (orange bar) at each resolution level. Notably, the training time costs of the deep learning model remain relatively stable across different iterations, whereas the label inference process exhibits a strong dependency on the number of ground atoms. In particular, the label inference module requires the most significant computational resources when the number of ground atoms is large.

To illustrate, in dataset 2, the label inference at the finest resolution consumed approximately $12.5$ hours without implementing a selective grounding process. However, when adopting the uncertainty-guided grounding strategy, the model achieved a considerable time-saving factor of $6.3$ and $5.0$ times compared to grounding all atoms for datasets 1 and 2, respectively. This stark difference in computational time underlines the necessity and effectiveness of our proposed uncertainty-guided hierarchical label inference in the context of large-scale spatial data.

%% file: main.bbl

\begin{thebibliography}{37}


\ifx \showCODEN    \undefined \def \showCODEN     #1{\unskip}     \fi
\ifx \showDOI      \undefined \def \showDOI       #1{#1}\fi
\ifx \showISBNx    \undefined \def \showISBNx     #1{\unskip}     \fi
\ifx \showISBNxiii \undefined \def \showISBNxiii  #1{\unskip}     \fi
\ifx \showISSN     \undefined \def \showISSN      #1{\unskip}     \fi
\ifx \showLCCN     \undefined \def \showLCCN      #1{\unskip}     \fi
\ifx \shownote     \undefined \def \shownote      #1{#1}          \fi
\ifx \showarticletitle \undefined \def \showarticletitle #1{#1}   \fi
\ifx \showURL      \undefined \def \showURL       {\relax}        \fi
\providecommand\bibfield[2]{#2}
\providecommand\bibinfo[2]{#2}
\providecommand\natexlab[1]{#1}
\providecommand\showeprint[2][]{arXiv:#2}

\bibitem[\protect\citeauthoryear{Bach, Broecheler, Huang, and Getoor}{Bach et~al\mbox{.}}{2017}]%
        {bach2017hinge}
\bibfield{author}{\bibinfo{person}{Stephen~H Bach}, \bibinfo{person}{Matthias Broecheler}, \bibinfo{person}{Bert Huang}, {and} \bibinfo{person}{Lise Getoor}.} \bibinfo{year}{2017}\natexlab{}.
\newblock \showarticletitle{Hinge-Loss Markov Random Fields and Probabilistic Soft Logic}.
\newblock \bibinfo{journal}{\emph{Journal of Machine Learning Research}}  \bibinfo{volume}{18} (\bibinfo{year}{2017}), \bibinfo{pages}{1--67}.
\newblock


\bibitem[\protect\citeauthoryear{Cai, Ding, Chen, Du, and Liu}{Cai et~al\mbox{.}}{2022}]%
        {cai2022mitigating}
\bibfield{author}{\bibinfo{person}{Bibo Cai}, \bibinfo{person}{Xiao Ding}, \bibinfo{person}{Bowen Chen}, \bibinfo{person}{Li Du}, {and} \bibinfo{person}{Ting Liu}.} \bibinfo{year}{2022}\natexlab{}.
\newblock \showarticletitle{Mitigating Reporting Bias in Semi-supervised Temporal Commonsense Inference with Probabilistic Soft Logic}. In \bibinfo{booktitle}{\emph{Proceedings of the AAAI Conference on Artificial Intelligence}}, Vol.~\bibinfo{volume}{36}. \bibinfo{pages}{10454--10462}.
\newblock


\bibitem[\protect\citeauthoryear{Dai, Xu, Yu, and Zhou}{Dai et~al\mbox{.}}{2019}]%
        {dai2019bridging}
\bibfield{author}{\bibinfo{person}{Wang-Zhou Dai}, \bibinfo{person}{Qiuling Xu}, \bibinfo{person}{Yang Yu}, {and} \bibinfo{person}{Zhi-Hua Zhou}.} \bibinfo{year}{2019}\natexlab{}.
\newblock \showarticletitle{Bridging machine learning and logical reasoning by abductive learning}.
\newblock \bibinfo{journal}{\emph{Advances in Neural Information Processing Systems}}  \bibinfo{volume}{32} (\bibinfo{year}{2019}).
\newblock


\bibitem[\protect\citeauthoryear{Diligenti, Gori, and Sacca}{Diligenti et~al\mbox{.}}{2017}]%
        {diligenti2017semantic}
\bibfield{author}{\bibinfo{person}{Michelangelo Diligenti}, \bibinfo{person}{Marco Gori}, {and} \bibinfo{person}{Claudio Sacca}.} \bibinfo{year}{2017}\natexlab{}.
\newblock \showarticletitle{Semantic-based regularization for learning and inference}.
\newblock \bibinfo{journal}{\emph{Artificial Intelligence}}  \bibinfo{volume}{244} (\bibinfo{year}{2017}), \bibinfo{pages}{143--165}.
\newblock


\bibitem[\protect\citeauthoryear{Donadello, Serafini, and d'Avila Garcez}{Donadello et~al\mbox{.}}{2017}]%
        {donadello2017logic}
\bibfield{author}{\bibinfo{person}{I Donadello}, \bibinfo{person}{L Serafini}, {and} \bibinfo{person}{AS d'Avila Garcez}.} \bibinfo{year}{2017}\natexlab{}.
\newblock \showarticletitle{Logic tensor networks for semantic image interpretation}. In \bibinfo{booktitle}{\emph{IJCAI International Joint Conference on Artificial Intelligence}}. IJCAI, \bibinfo{pages}{1596--1602}.
\newblock


\bibitem[\protect\citeauthoryear{Fabrizio, Farina, and De~Maio}{Fabrizio et~al\mbox{.}}{2006}]%
        {fabrizio2006knowledge}
\bibfield{author}{\bibinfo{person}{Giuseppe Fabrizio}, \bibinfo{person}{Alfonso Farina}, {and} \bibinfo{person}{Antonio De~Maio}.} \bibinfo{year}{2006}\natexlab{}.
\newblock \showarticletitle{Knowledge-based adaptive processing for ship detection in OTH Radar}. In \bibinfo{booktitle}{\emph{2006 International Radar Symposium}}. IEEE, \bibinfo{pages}{1--5}.
\newblock


\bibitem[\protect\citeauthoryear{Foulds and Frank}{Foulds and Frank}{2010}]%
        {foulds2010review}
\bibfield{author}{\bibinfo{person}{James Foulds} {and} \bibinfo{person}{Eibe Frank}.} \bibinfo{year}{2010}\natexlab{}.
\newblock \showarticletitle{A review of multi-instance learning assumptions}.
\newblock \bibinfo{journal}{\emph{The knowledge engineering review}} \bibinfo{volume}{25}, \bibinfo{number}{1} (\bibinfo{year}{2010}), \bibinfo{pages}{1--25}.
\newblock


\bibitem[\protect\citeauthoryear{Gao, He, Sun, Jia, and Zhang}{Gao et~al\mbox{.}}{2019}]%
        {gao2019incorporating}
\bibfield{author}{\bibinfo{person}{Lianru Gao}, \bibinfo{person}{Yiqun He}, \bibinfo{person}{Xu Sun}, \bibinfo{person}{Xiuping Jia}, {and} \bibinfo{person}{Bing Zhang}.} \bibinfo{year}{2019}\natexlab{}.
\newblock \showarticletitle{Incorporating negative sample training for ship detection based on deep learning}.
\newblock \bibinfo{journal}{\emph{Sensors}} \bibinfo{volume}{19}, \bibinfo{number}{3} (\bibinfo{year}{2019}), \bibinfo{pages}{684}.
\newblock


\bibitem[\protect\citeauthoryear{Garcez, Bader, Bowman, Lamb, de~Penning, Illuminoo, Poon, and Zaverucha}{Garcez et~al\mbox{.}}{2022}]%
        {garcez2022neural}
\bibfield{author}{\bibinfo{person}{Artur~d’Avila Garcez}, \bibinfo{person}{Sebastian Bader}, \bibinfo{person}{Howard Bowman}, \bibinfo{person}{Luis~C Lamb}, \bibinfo{person}{Leo de Penning}, \bibinfo{person}{BV Illuminoo}, \bibinfo{person}{Hoifung Poon}, {and} \bibinfo{person}{COPPE~Gerson Zaverucha}.} \bibinfo{year}{2022}\natexlab{}.
\newblock \showarticletitle{Neural-symbolic learning and reasoning: a survey and interpretation}.
\newblock \bibinfo{journal}{\emph{Neuro-Symbolic Artificial Intelligence: The State of the Art}} \bibinfo{volume}{342}, \bibinfo{number}{1} (\bibinfo{year}{2022}).
\newblock


\bibitem[\protect\citeauthoryear{Harmon, Marconi, Weinstein, Graves, Wang, Zare, Bohlman, Singh, and White}{Harmon et~al\mbox{.}}{2022}]%
        {harmon2022injecting}
\bibfield{author}{\bibinfo{person}{Ira Harmon}, \bibinfo{person}{Sergio Marconi}, \bibinfo{person}{Ben Weinstein}, \bibinfo{person}{Sarah Graves}, \bibinfo{person}{Daisy~Zhe Wang}, \bibinfo{person}{Alina Zare}, \bibinfo{person}{Stephanie Bohlman}, \bibinfo{person}{Aditya Singh}, {and} \bibinfo{person}{Ethan White}.} \bibinfo{year}{2022}\natexlab{}.
\newblock \showarticletitle{Injecting Domain Knowledge Into Deep Neural Networks for Tree Crown Delineation}.
\newblock \bibinfo{journal}{\emph{IEEE Transactions on Geoscience and Remote Sensing}}  \bibinfo{volume}{60} (\bibinfo{year}{2022}), \bibinfo{pages}{1--19}.
\newblock


\bibitem[\protect\citeauthoryear{He and Jiang}{He and Jiang}{2023}]%
        {he2023survey}
\bibfield{author}{\bibinfo{person}{Wenchong He} {and} \bibinfo{person}{Zhe Jiang}.} \bibinfo{year}{2023}\natexlab{}.
\newblock \showarticletitle{A Survey on Uncertainty Quantification Methods for Deep Neural Networks: An Uncertainty Source Perspective}.
\newblock \bibinfo{journal}{\emph{arXiv preprint arXiv:2302.13425}} (\bibinfo{year}{2023}).
\newblock


\bibitem[\protect\citeauthoryear{He, Jiang, Kriby, Xie, Jia, Yan, and Zhou}{He et~al\mbox{.}}{2022a}]%
        {he2022quantifying}
\bibfield{author}{\bibinfo{person}{Wenchong He}, \bibinfo{person}{Zhe Jiang}, \bibinfo{person}{Marcus Kriby}, \bibinfo{person}{Yiqun Xie}, \bibinfo{person}{Xiaowei Jia}, \bibinfo{person}{Da Yan}, {and} \bibinfo{person}{Yang Zhou}.} \bibinfo{year}{2022}\natexlab{a}.
\newblock \showarticletitle{Quantifying and Reducing Registration Uncertainty of Spatial Vector Labels on Earth Imagery}. In \bibinfo{booktitle}{\emph{Proceedings of the 28th ACM SIGKDD Conference on Knowledge Discovery and Data Mining}}. \bibinfo{pages}{554--564}.
\newblock


\bibitem[\protect\citeauthoryear{He, Sainju, Jiang, Yan, and Zhou}{He et~al\mbox{.}}{2022b}]%
        {he2022earth}
\bibfield{author}{\bibinfo{person}{Wenchong He}, \bibinfo{person}{Arpan~Man Sainju}, \bibinfo{person}{Zhe Jiang}, \bibinfo{person}{Da Yan}, {and} \bibinfo{person}{Yang Zhou}.} \bibinfo{year}{2022}\natexlab{b}.
\newblock \showarticletitle{Earth Imagery Segmentation on Terrain Surface with Limited Training Labels: A Semi-supervised Approach based on Physics-Guided Graph Co-Training}.
\newblock \bibinfo{journal}{\emph{ACM Transactions on Intelligent Systems and Technology (TIST)}} \bibinfo{volume}{13}, \bibinfo{number}{2} (\bibinfo{year}{2022}), \bibinfo{pages}{1--22}.
\newblock


\bibitem[\protect\citeauthoryear{Hu, Ma, Liu, Hovy, and Xing}{Hu et~al\mbox{.}}{2016}]%
        {hu2016harnessing}
\bibfield{author}{\bibinfo{person}{Zhiting Hu}, \bibinfo{person}{Xuezhe Ma}, \bibinfo{person}{Zhengzhong Liu}, \bibinfo{person}{Eduard Hovy}, {and} \bibinfo{person}{Eric Xing}.} \bibinfo{year}{2016}\natexlab{}.
\newblock \showarticletitle{Harnessing Deep Neural Networks with Logic Rules}. In \bibinfo{booktitle}{\emph{Proceedings of the 54th Annual Meeting of the Association for Computational Linguistics}}. \bibinfo{pages}{2410--2420}.
\newblock


\bibitem[\protect\citeauthoryear{Huang, Dai, Yang, Cai, Cheng, Huang, Li, and Zhou}{Huang et~al\mbox{.}}{2020}]%
        {huang2020semi}
\bibfield{author}{\bibinfo{person}{Yu-Xuan Huang}, \bibinfo{person}{Wang-Zhou Dai}, \bibinfo{person}{Jian Yang}, \bibinfo{person}{Le-Wen Cai}, \bibinfo{person}{Shaofen Cheng}, \bibinfo{person}{Ruizhang Huang}, \bibinfo{person}{Yu-Feng Li}, {and} \bibinfo{person}{Zhi-Hua Zhou}.} \bibinfo{year}{2020}\natexlab{}.
\newblock \showarticletitle{Semi-supervised abductive learning and its application to theft judicial sentencing}. In \bibinfo{booktitle}{\emph{2020 IEEE international conference on data mining (ICDM)}}. IEEE, \bibinfo{pages}{1070--1075}.
\newblock


\bibitem[\protect\citeauthoryear{Jean, Wang, Samar, Azzari, Lobell, and Ermon}{Jean et~al\mbox{.}}{2019}]%
        {jean2019tile2vec}
\bibfield{author}{\bibinfo{person}{Neal Jean}, \bibinfo{person}{Sherrie Wang}, \bibinfo{person}{Anshul Samar}, \bibinfo{person}{George Azzari}, \bibinfo{person}{David Lobell}, {and} \bibinfo{person}{Stefano Ermon}.} \bibinfo{year}{2019}\natexlab{}.
\newblock \showarticletitle{Tile2vec: Unsupervised representation learning for spatially distributed data}. In \bibinfo{booktitle}{\emph{Proceedings of the AAAI Conference on Artificial Intelligence}}, Vol.~\bibinfo{volume}{33}. \bibinfo{pages}{3967--3974}.
\newblock


\bibitem[\protect\citeauthoryear{Jiang, Xie, and Sainju}{Jiang et~al\mbox{.}}{2019}]%
        {jiang2019geographical}
\bibfield{author}{\bibinfo{person}{Zhe Jiang}, \bibinfo{person}{Miao Xie}, {and} \bibinfo{person}{Arpan~Man Sainju}.} \bibinfo{year}{2019}\natexlab{}.
\newblock \showarticletitle{Geographical hidden Markov tree}.
\newblock \bibinfo{journal}{\emph{IEEE Transactions on Knowledge and Data Engineering}} \bibinfo{volume}{33}, \bibinfo{number}{2} (\bibinfo{year}{2019}), \bibinfo{pages}{506--520}.
\newblock


\bibitem[\protect\citeauthoryear{Karpatne, Kannan, and Kumar}{Karpatne et~al\mbox{.}}{2022}]%
        {karpatne2022knowledge}
\bibfield{author}{\bibinfo{person}{Anuj Karpatne}, \bibinfo{person}{Ramakrishnan Kannan}, {and} \bibinfo{person}{Vipin Kumar}.} \bibinfo{year}{2022}\natexlab{}.
\newblock \bibinfo{booktitle}{\emph{Knowledge Guided Machine Learning: Accelerating Discovery Using Scientific Knowledge and Data}}.
\newblock \bibinfo{publisher}{CRC Press}.
\newblock


\bibitem[\protect\citeauthoryear{Kimmig, Bach, Broecheler, Huang, and Getoor}{Kimmig et~al\mbox{.}}{2012}]%
        {kimmig2012short}
\bibfield{author}{\bibinfo{person}{Angelika Kimmig}, \bibinfo{person}{Stephen Bach}, \bibinfo{person}{Matthias Broecheler}, \bibinfo{person}{Bert Huang}, {and} \bibinfo{person}{Lise Getoor}.} \bibinfo{year}{2012}\natexlab{}.
\newblock \showarticletitle{A short introduction to probabilistic soft logic}. In \bibinfo{booktitle}{\emph{Proceedings of the NIPS workshop on probabilistic programming: foundations and applications}}. \bibinfo{pages}{1--4}.
\newblock


\bibitem[\protect\citeauthoryear{Krishnan and Tickoo}{Krishnan and Tickoo}{2020}]%
        {krishnan2020improving}
\bibfield{author}{\bibinfo{person}{Ranganath Krishnan} {and} \bibinfo{person}{Omesh Tickoo}.} \bibinfo{year}{2020}\natexlab{}.
\newblock \showarticletitle{Improving model calibration with accuracy versus uncertainty optimization}.
\newblock \bibinfo{journal}{\emph{Advances in Neural Information Processing Systems}}  \bibinfo{volume}{33} (\bibinfo{year}{2020}), \bibinfo{pages}{18237--18248}.
\newblock


\bibitem[\protect\citeauthoryear{Li, Huang, Hong, Chen, Wu, and Zhu}{Li et~al\mbox{.}}{2020}]%
        {li2020closed}
\bibfield{author}{\bibinfo{person}{Qing Li}, \bibinfo{person}{Siyuan Huang}, \bibinfo{person}{Yining Hong}, \bibinfo{person}{Yixin Chen}, \bibinfo{person}{Ying~Nian Wu}, {and} \bibinfo{person}{Song-Chun Zhu}.} \bibinfo{year}{2020}\natexlab{}.
\newblock \showarticletitle{Closed loop neural-symbolic learning via integrating neural perception, grammar parsing, and symbolic reasoning}. In \bibinfo{booktitle}{\emph{International Conference on Machine Learning}}. PMLR, \bibinfo{pages}{5884--5894}.
\newblock


\bibitem[\protect\citeauthoryear{Manhaeve, Dumancic, Kimmig, Demeester, and De~Raedt}{Manhaeve et~al\mbox{.}}{2018}]%
        {manhaeve2018deepproblog}
\bibfield{author}{\bibinfo{person}{Robin Manhaeve}, \bibinfo{person}{Sebastijan Dumancic}, \bibinfo{person}{Angelika Kimmig}, \bibinfo{person}{Thomas Demeester}, {and} \bibinfo{person}{Luc De~Raedt}.} \bibinfo{year}{2018}\natexlab{}.
\newblock \showarticletitle{Deepproblog: Neural probabilistic logic programming}.
\newblock \bibinfo{journal}{\emph{advances in neural information processing systems}}  \bibinfo{volume}{31} (\bibinfo{year}{2018}).
\newblock


\bibitem[\protect\citeauthoryear{Marra and Ku{\v{z}}elka}{Marra and Ku{\v{z}}elka}{2021}]%
        {marra2021neural}
\bibfield{author}{\bibinfo{person}{Giuseppe Marra} {and} \bibinfo{person}{Ond{\v{r}}ej Ku{\v{z}}elka}.} \bibinfo{year}{2021}\natexlab{}.
\newblock \showarticletitle{Neural markov logic networks}. In \bibinfo{booktitle}{\emph{Uncertainty in Artificial Intelligence}}. PMLR, \bibinfo{pages}{908--917}.
\newblock


\bibitem[\protect\citeauthoryear{Qiao, He, Cheng, Li, Zhao, Zhao, and Tian}{Qiao et~al\mbox{.}}{2023}]%
        {qiao2023kstage}
\bibfield{author}{\bibinfo{person}{Mengjia Qiao}, \bibinfo{person}{Xiaohui He}, \bibinfo{person}{Xijie Cheng}, \bibinfo{person}{Panle Li}, \bibinfo{person}{Qianbo Zhao}, \bibinfo{person}{Chenlu Zhao}, {and} \bibinfo{person}{Zhihui Tian}.} \bibinfo{year}{2023}\natexlab{}.
\newblock \showarticletitle{KSTAGE: A knowledge-guided spatial-temporal attention graph learning network for crop yield prediction}.
\newblock \bibinfo{journal}{\emph{Information Sciences}}  \bibinfo{volume}{619} (\bibinfo{year}{2023}), \bibinfo{pages}{19--37}.
\newblock


\bibitem[\protect\citeauthoryear{Qu and Tang}{Qu and Tang}{2019}]%
        {qu2019probabilistic}
\bibfield{author}{\bibinfo{person}{Meng Qu} {and} \bibinfo{person}{Jian Tang}.} \bibinfo{year}{2019}\natexlab{}.
\newblock \showarticletitle{Probabilistic logic neural networks for reasoning}.
\newblock \bibinfo{journal}{\emph{Advances in neural information processing systems}}  \bibinfo{volume}{32} (\bibinfo{year}{2019}).
\newblock


\bibitem[\protect\citeauthoryear{Richardson and Domingos}{Richardson and Domingos}{2006}]%
        {richardson2006markov}
\bibfield{author}{\bibinfo{person}{Matthew Richardson} {and} \bibinfo{person}{Pedro Domingos}.} \bibinfo{year}{2006}\natexlab{}.
\newblock \showarticletitle{Markov logic networks}.
\newblock \bibinfo{journal}{\emph{Machine learning}}  \bibinfo{volume}{62} (\bibinfo{year}{2006}), \bibinfo{pages}{107--136}.
\newblock


\bibitem[\protect\citeauthoryear{Ronneberger, Fischer, and Brox}{Ronneberger et~al\mbox{.}}{2015}]%
        {ronneberger2015u}
\bibfield{author}{\bibinfo{person}{Olaf Ronneberger}, \bibinfo{person}{Philipp Fischer}, {and} \bibinfo{person}{Thomas Brox}.} \bibinfo{year}{2015}\natexlab{}.
\newblock \showarticletitle{U-Net: Convolutional Networks for Biomedical Image Segmentation}. In \bibinfo{booktitle}{\emph{Medical Image Computing and Computer-Assisted Intervention (MICCAI)}}.
\newblock


\bibitem[\protect\citeauthoryear{Ru{\ss}wurm and Korner}{Ru{\ss}wurm and Korner}{2017}]%
        {russwurm2017temporal}
\bibfield{author}{\bibinfo{person}{Marc Ru{\ss}wurm} {and} \bibinfo{person}{Marco Korner}.} \bibinfo{year}{2017}\natexlab{}.
\newblock \showarticletitle{Temporal vegetation modelling using long short-term memory networks for crop identification from medium-resolution multi-spectral satellite images}. In \bibinfo{booktitle}{\emph{Proceedings of the IEEE conference on computer vision and pattern recognition workshops}}. \bibinfo{pages}{11--19}.
\newblock


\bibitem[\protect\citeauthoryear{Sainju, He, and Jiang}{Sainju et~al\mbox{.}}{2020}]%
        {sainju2020hidden}
\bibfield{author}{\bibinfo{person}{Arpan~Man Sainju}, \bibinfo{person}{Wenchong He}, {and} \bibinfo{person}{Zhe Jiang}.} \bibinfo{year}{2020}\natexlab{}.
\newblock \showarticletitle{A hidden markov contour tree model for spatial structured prediction}.
\newblock \bibinfo{journal}{\emph{IEEE Transactions on Knowledge and Data Engineering}} \bibinfo{volume}{34}, \bibinfo{number}{4} (\bibinfo{year}{2020}), \bibinfo{pages}{1530--1543}.
\newblock


\bibitem[\protect\citeauthoryear{Tian, Li, Chen, Xiao, He, and Jin}{Tian et~al\mbox{.}}{2022}]%
        {tian2022weakly}
\bibfield{author}{\bibinfo{person}{Jidong Tian}, \bibinfo{person}{Yitian Li}, \bibinfo{person}{Wenqing Chen}, \bibinfo{person}{Liqiang Xiao}, \bibinfo{person}{Hao He}, {and} \bibinfo{person}{Yaohui Jin}.} \bibinfo{year}{2022}\natexlab{}.
\newblock \showarticletitle{Weakly Supervised Neural Symbolic Learning for Cognitive Tasks}. In \bibinfo{booktitle}{\emph{Proceedings of the AAAI Conference on Artificial Intelligence}}, Vol.~\bibinfo{volume}{36}. \bibinfo{pages}{5888--5896}.
\newblock


\bibitem[\protect\citeauthoryear{Weber, Minervini, M{\"u}nchmeyer, Leser, and Rockt{\"a}schl}{Weber et~al\mbox{.}}{2019}]%
        {weber2019nlprolog}
\bibfield{author}{\bibinfo{person}{L Weber}, \bibinfo{person}{P Minervini}, \bibinfo{person}{J M{\"u}nchmeyer}, \bibinfo{person}{U Leser}, {and} \bibinfo{person}{T Rockt{\"a}schl}.} \bibinfo{year}{2019}\natexlab{}.
\newblock \showarticletitle{NLProlog: Reasoning with Weak Unification for Question Answering in Natural Language}. In \bibinfo{booktitle}{\emph{57th Annual Meeting of the Association for Computational Linguistics}}. Association for Computational Linguistics, \bibinfo{pages}{6151--6161}.
\newblock


\bibitem[\protect\citeauthoryear{Xie, Jiang, and Sainju}{Xie et~al\mbox{.}}{2018}]%
        {xie2018geographical}
\bibfield{author}{\bibinfo{person}{Miao Xie}, \bibinfo{person}{Zhe Jiang}, {and} \bibinfo{person}{Arpan~Man Sainju}.} \bibinfo{year}{2018}\natexlab{}.
\newblock \showarticletitle{Geographical hidden markov tree for flood extent mapping}. In \bibinfo{booktitle}{\emph{Proceedings of the 24th ACM SIGKDD International Conference on Knowledge Discovery \& Data Mining}}. \bibinfo{pages}{2545--2554}.
\newblock


\bibitem[\protect\citeauthoryear{Xie, Xu, Kankanhalli, Meel, and Soh}{Xie et~al\mbox{.}}{2019}]%
        {xie2019embedding}
\bibfield{author}{\bibinfo{person}{Yaqi Xie}, \bibinfo{person}{Ziwei Xu}, \bibinfo{person}{Mohan~S Kankanhalli}, \bibinfo{person}{Kuldeep~S Meel}, {and} \bibinfo{person}{Harold Soh}.} \bibinfo{year}{2019}\natexlab{}.
\newblock \showarticletitle{Embedding symbolic knowledge into deep networks}.
\newblock \bibinfo{journal}{\emph{Advances in neural information processing systems}}  \bibinfo{volume}{32} (\bibinfo{year}{2019}).
\newblock


\bibitem[\protect\citeauthoryear{Xu, Zhang, Friedman, Liang, and Broeck}{Xu et~al\mbox{.}}{2018}]%
        {xu2018semantic}
\bibfield{author}{\bibinfo{person}{Jingyi Xu}, \bibinfo{person}{Zilu Zhang}, \bibinfo{person}{Tal Friedman}, \bibinfo{person}{Yitao Liang}, {and} \bibinfo{person}{Guy Broeck}.} \bibinfo{year}{2018}\natexlab{}.
\newblock \showarticletitle{A semantic loss function for deep learning with symbolic knowledge}. In \bibinfo{booktitle}{\emph{International conference on machine learning}}. PMLR, \bibinfo{pages}{5502--5511}.
\newblock


\bibitem[\protect\citeauthoryear{Yu, Yang, Wei, Li, and Pan}{Yu et~al\mbox{.}}{2022}]%
        {yu2022probabilistic}
\bibfield{author}{\bibinfo{person}{Dongran Yu}, \bibinfo{person}{Bo Yang}, \bibinfo{person}{Qianhao Wei}, \bibinfo{person}{Anchen Li}, {and} \bibinfo{person}{Shirui Pan}.} \bibinfo{year}{2022}\natexlab{}.
\newblock \showarticletitle{A probabilistic graphical model based on neural-symbolic reasoning for visual relationship detection}. In \bibinfo{booktitle}{\emph{Proceedings of the IEEE/CVF Conference on Computer Vision and Pattern Recognition}}. \bibinfo{pages}{10609--10618}.
\newblock


\bibitem[\protect\citeauthoryear{Zhang, Chen, Yang, Ramamurthy, Li, Qi, and Song}{Zhang et~al\mbox{.}}{2019}]%
        {zhang2019efficient}
\bibfield{author}{\bibinfo{person}{Yuyu Zhang}, \bibinfo{person}{Xinshi Chen}, \bibinfo{person}{Yuan Yang}, \bibinfo{person}{Arun Ramamurthy}, \bibinfo{person}{Bo Li}, \bibinfo{person}{Yuan Qi}, {and} \bibinfo{person}{Le Song}.} \bibinfo{year}{2019}\natexlab{}.
\newblock \showarticletitle{Efficient Probabilistic Logic Reasoning with Graph Neural Networks}. In \bibinfo{booktitle}{\emph{International Conference on Learning Representations}}.
\newblock


\bibitem[\protect\citeauthoryear{Zhou, Yan, Han, Caufield, Chang, Sun, Ping, and Wang}{Zhou et~al\mbox{.}}{2021}]%
        {zhou2021clinical}
\bibfield{author}{\bibinfo{person}{Yichao Zhou}, \bibinfo{person}{Yu Yan}, \bibinfo{person}{Rujun Han}, \bibinfo{person}{J~Harry Caufield}, \bibinfo{person}{Kai-Wei Chang}, \bibinfo{person}{Yizhou Sun}, \bibinfo{person}{Peipei Ping}, {and} \bibinfo{person}{Wei Wang}.} \bibinfo{year}{2021}\natexlab{}.
\newblock \showarticletitle{Clinical temporal relation extraction with probabilistic soft logic regularization and global inference}. In \bibinfo{booktitle}{\emph{Proceedings of the AAAI Conference on Artificial Intelligence}}, Vol.~\bibinfo{volume}{35}. \bibinfo{pages}{14647--14655}.
\newblock


\end{thebibliography}
